\documentclass{article} 
\usepackage{nips15submit_e,times}
\usepackage{hyperref}
\usepackage{url}

\usepackage{times}
\usepackage{helvet}
\usepackage{courier}

\usepackage{amsthm,thmtools}
\usepackage{mnick}

\usepackage{graphicx} 

\usepackage[numbers]{natbib}

\usepackage{algorithm}
\usepackage{algorithmic}

\usepackage{hyperref}

\usepackage{enumitem}
\usepackage{booktabs}
\usepackage{multicol}

\title{
Learning with Memory Embeddings
}

\author{
Volker Tresp, Crist\'{o}bal  Esteban, Yinchong Yang, \\ \textbf{{Stephan Baier and Denis Krompa{\ss}}}  \\
Siemens AG and Ludwig Maximilian University  of Munich, Germany
\texttt{}}

\nipsfinalcopy 

\begin{document}

\maketitle

\begin{abstract}

Embedding learning, a.k.a. representation learning,  has been shown to be able to model large-scale semantic knowledge graphs. A  key concept is a mapping of the knowledge graph to  a tensor representation whose entries are predicted by models using latent representations of generalized entities.  Latent variable models are well suited to deal with the high dimensionality and sparsity of typical knowledge graphs. In recent publications the embedding models were extended to also consider time evolutions, time patterns and subsymbolic representations. In this paper  we  map   embedding models, which were developed purely as solutions to technical problems for modelling temporal knowledge graphs, to various cognitive memory functions, in particular to semantic and concept memory, episodic memory, sensory memory, short-term memory, and working memory. We discuss  learning, query answering,  the path from sensory input to semantic decoding,  and the relationship between episodic memory and semantic memory.
We introduce a number of hypotheses on human memory that can be derived from the developed mathematical models.
There are  four main hypotheses. The first one is that  semantic memory is described as triples and that  episodic memory is described as triples in time. A second main hypothesis is that generalized entities have  unique latent representations which are shared across memory functions and that are the basis for prediction, decision support and other functionalities executed by working memory (tensor memory hypothesis). A third main hypothesis is that the  latent representation for a time $t$, which summarizes all sensory information available at time $t$, is the basis for episodic memory. Finally, our proposed model suggests that semantic memory and episodic memory depend on each other: Episodic decoding depends on semantic memory and  semantic memory is developed as a long term store of episodic memory.     On the other hand there is also a certain independence: the pure storage of episodic memory does not depend on semantic memory and semantic memory can be acquired even without a functioning episodic memory.  The same relationships between semantic and episodic memories can be found  in the human brain.


\end{abstract}

\section{Introduction}

Embedding learning, a.k.a. representation learning, is an essential ingredient of successful
natural language models and deep architectures~\cite{schutze1993word,bengio1,bengio2012deep,bengio2013representation,lecun2015deep,goodfellow2015deep} and has been the basis for modelling large-scale semantic knowledge graphs~\cite{paccanaro2001learning,tresp_materializing_2009,nickel_three-way_2011,bordes2011learning,bordes_translating_2013,socher_reasoning_2013,dong_knowledge_2014,nickel2015,nickel2015holographic}.\footnote{Some authors make a distinction between latent representations, which are application specific, and embeddings, which are identical across applications and might represent universal properties of entities~\cite{rothe2015autoextend,Schutze2016}.  }
A  key concept is a mapping of the knowledge graph to  a tensor representation whose entries are predicted by models using latent representations of generalized entities. Latent variable models are well suited to deal with the high dimensionality and sparsity of typical knowledge graphs.
In recent publications the embedding models were extended to also consider temporal evolutions, time patterns and subsymbolic representations~\cite{Crist2015,EstebanEvent}.
These extended models were  used successfully  to predict clinical events like procedures, lab measurements, and diagnoses.
In this paper,  we attempt to map  these embedding models, which were developed purely as solutions to technical problems, to
 various cognitive memory functions.
  Our approach  follows the tradition of
latent semantic analysis (LSA), which  is a classical  representation learning approach that on the one hand  has found a number of technical applications and on the other hand could be related to
cognitive semantic memories~\cite{landauer1998introduction,landauer1997solution,DBLP:journals/jasis/DeerwesterDLFH90}.

\begin{figure}
	\centering
 	\includegraphics[width=\columnwidth]{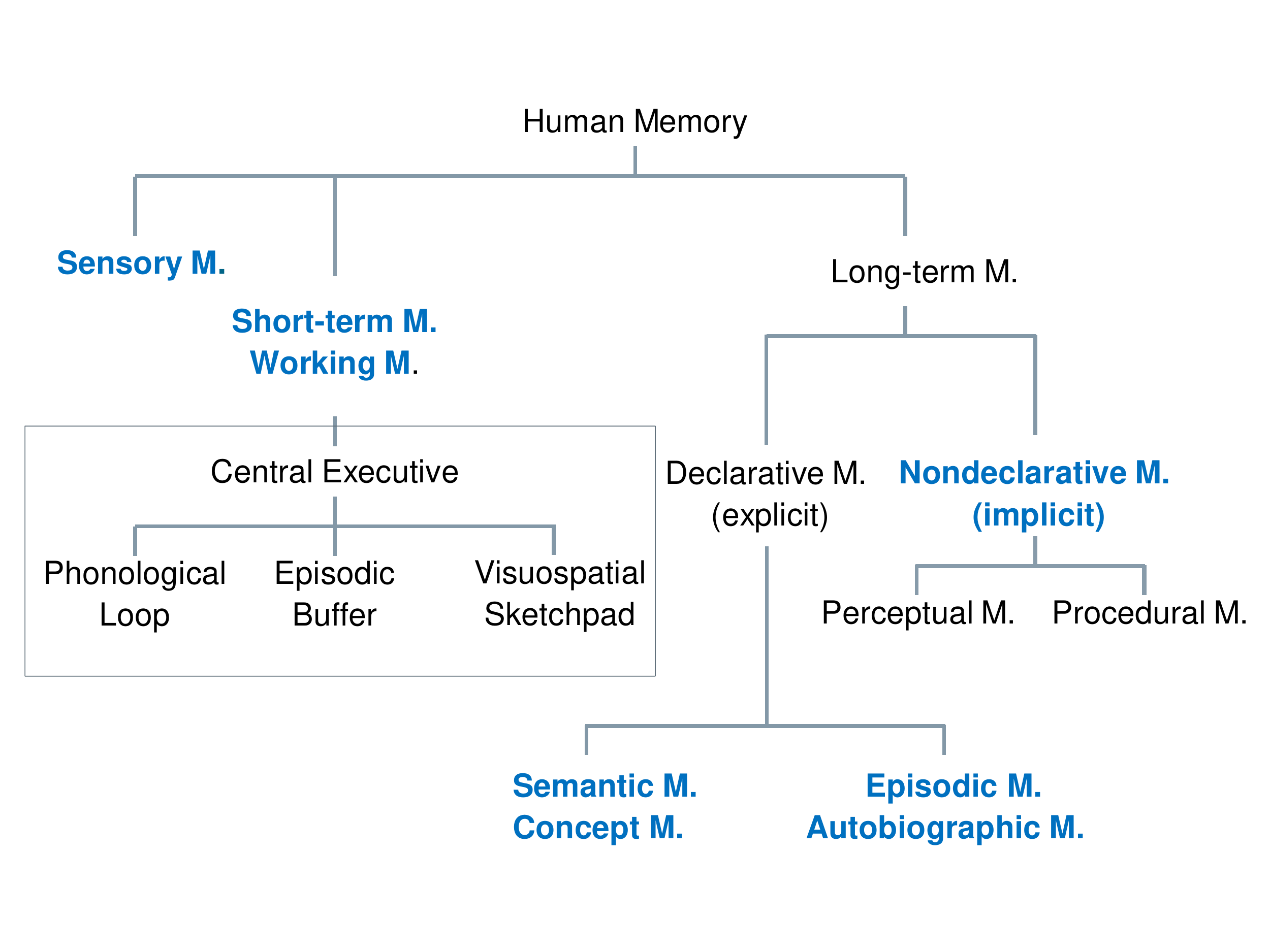}
	\caption{%
		Organization of human memory~\cite{gazzaniga2004cognitive,gluck2013learning}. In this paper,  we discuss the memory functions in blue.  Sensory memory, episodic memory and semantic memory will be discussed in most sections.
Autobiographic memory is the topic of Subsection~\ref{sec:auto}.
Working memory and short-term memory are discussed in  Sections~\ref{sec:predict} and Subsection~\ref{sec:wm2}.
	Compare Figures~\ref{fig:memHier} and~\ref{fig:NN-bio}.}
	\label{fig:mem}
\end{figure}

Cognitive memory functions are typically classified as   \emph{long-term},  \emph{short-term}, and \emph{sensory}  memory, where  long-term memory has the subcategories   \emph{declarative} memory and \emph{non-declarative} memory~\cite{ebbinghaus1885gedachtnis,atkinson1968human,squire1987memory,bartlett1995remembering,cowan2008differences,gazzaniga2004cognitive,gluck2013learning}. Figure~\ref{fig:mem} shows these main categories and finer subcategories and shows the role of working memory~\cite{baddeley1992working}.
There is evidence that these main cognitive categories
are partially dissociated from one another in the brain, as expressed in their differential sensitivity to brain damage~\cite{gazzaniga2004cognitive}. However,  there is also evidence indicating  that the different  memory functions are not mutually independent and support each other~\cite{jonides2008mind,greenberg2010interdependence}.

The paper is organized as follows.
 In the next section, we introduce the unique-representation hypothesis as the basis for exchanging information between different memory functions.
 We present the different tensor representations of the main memory functions and discuss offline learning of the models.
 In Section~\ref{sec:modelquery} we introduce different representations for the indicator mapping function used in the memory models  and in Section~\ref{sec:qa} we show  how likely triples can be generated from the  model using a simulated-annealing based sampling perspective.
 In Section~\ref{sec:senssem} we discuss the path from sensory input to a semantic representation of scene information and to long-term semantic and episodic memory.
In Section~\ref{sec:predict} we explain how the different memory representations form  the basis of a prediction system and relate this to working memory.
Section~\ref{sec:shared} represents the main results of this paper in form of  a discussion of a number of postulated hypotheses for human memory.
Section~\ref{sec:concl} contains our conclusions.

\begin{figure}
	\centering
	\includegraphics[width=\columnwidth]{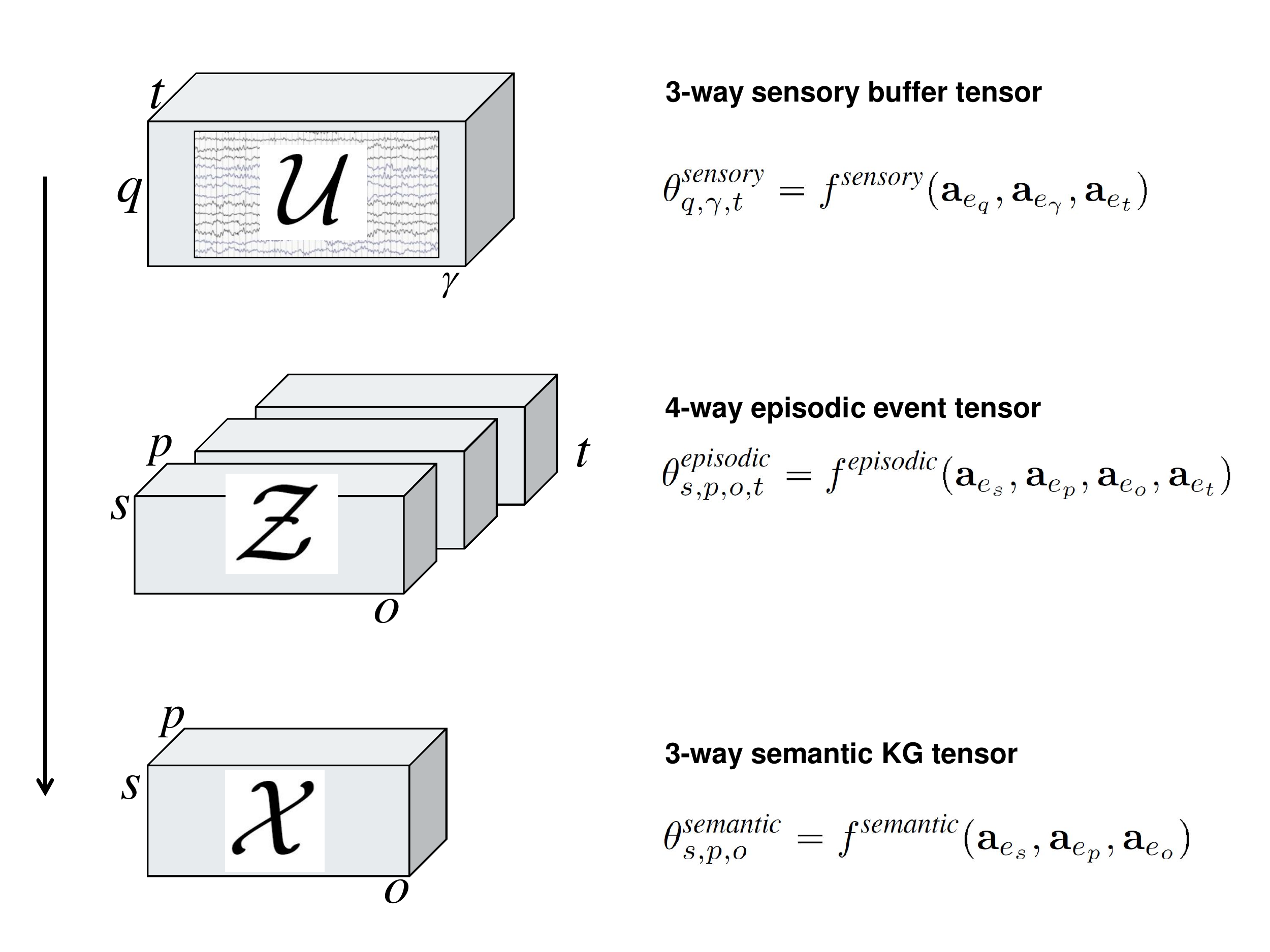}
	\caption{%
 The figure shows the different tensor memories and their models. On the top we see the sensory memory tensor $\mathcal{U}$ with dimensions sensory channel $q$, within buffer position $\gamma$,  and time  $t$. The time dimension is shared with the episodic event tensor  tensor $\mathcal{Z}$ with additional dimensions subject $s$, predicate $p$, and object $o$.  The latter three are shared with the semantic KG tensor $\mathcal{X}$. On the right side we show the indicator mapping functions, which are functions of latent representations of the involved generalized entities.
	}
	\label{fig:memHier}
\end{figure}

\section{Memories and Their Tensor Embeddings}
\label{sec:memtens}

\subsection{Unique-Representation Hypothesis}

In this section we discuss how the different memory functions can be coded as tensors and how inference and generalization can be achieved by coupled tensor decompositions.

We begin by considering declarative  memories.  The prime example of a declarative memory is the \textit{semantic memory}  which stores general world knowledge about  entities. Second, there is
\emph{concept memory} which stores information about the concepts in the world and their hierarchical organization.
{In contrast to the general setting in  machine learning, in this paper entities are the prime focus and concepts are of secondary interest.}
Finally,
 \emph{episodic memory}  stores   information of general and personal events~\cite{tulving1972episodic,tulving1985elements,tulving2002episodic,gazzaniga2004cognitive}.
 Whereas semantic memory concerns information we ``know'', episodic memory concerns information we ``remember''~\cite{gluck2013learning}.
 The portion of episodic memory that concerns an individual's life involving personal experiences  is called autobiographic memory.

Semantic memories and episodic memories are long-term memories. In contrast, we also consider sensory memory, which is the shortest-time   element of memory. It is the ability to retain impressions of sensory information after the original stimuli have ended~\cite{gazzaniga2004cognitive}.

Finally, working memory is the topic of   Section~\ref{sec:predict}. {Working memory uses  the other memories
  for tasks like prediction, decision support and other high-level functions. }

\begin{figure}
	\centering
	\includegraphics[width=0.7\columnwidth]{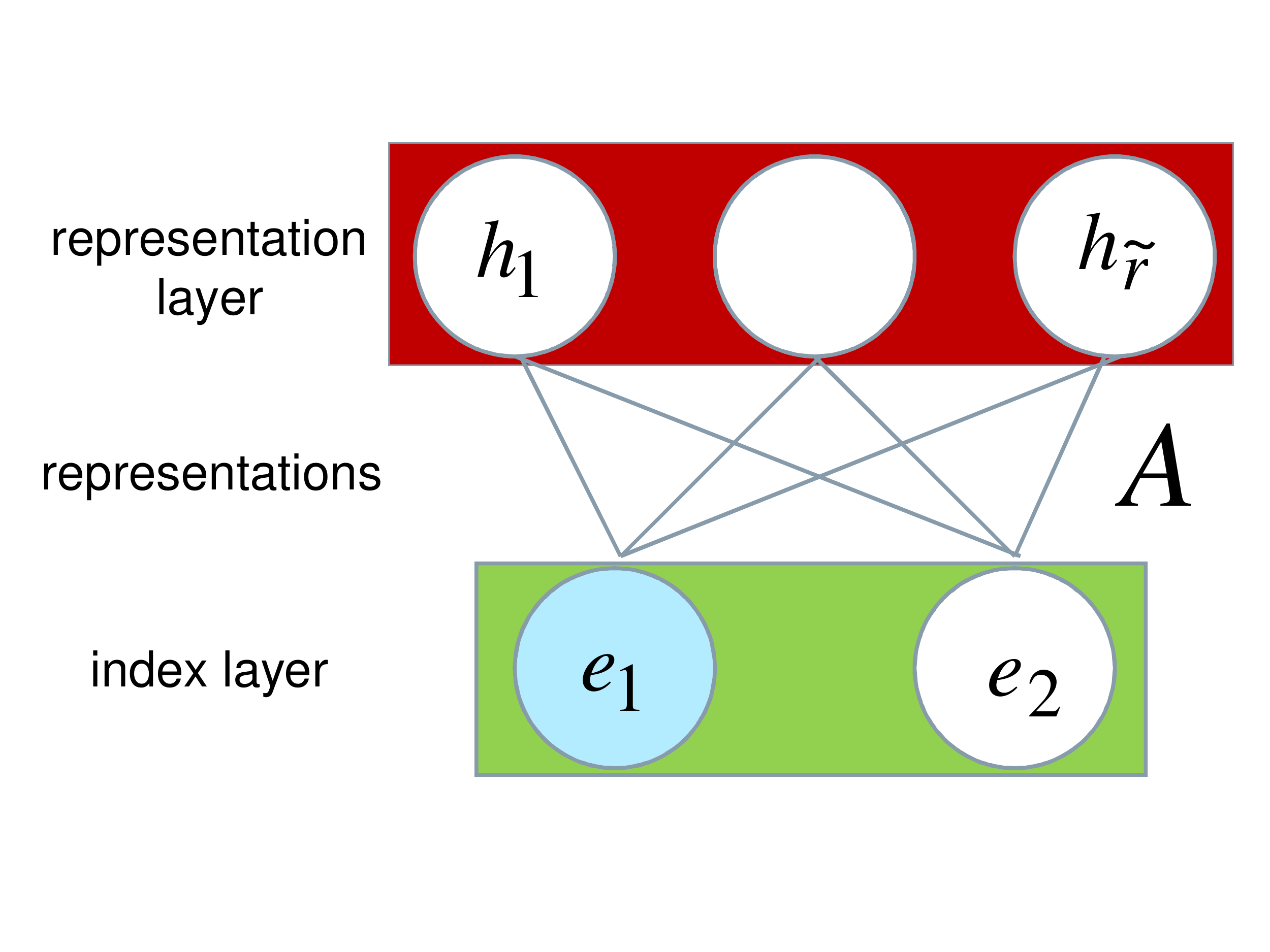}
	\caption{%
A graphical view of  the  unique-representation  hypothesis.  The model can operate bottom up and top down. In the first case, index neurons $e_i$ activate the representation layer via their latent representations, implemented as weight vectors. In the figure $e_1$ is active, all other neurons are inactive and the representation layer is activated with the pattern  $\mathbf{h} = \mathbf{a}_{e_1}$. In top-down operation, a representation layer can also activate index neurons.   The activation of neuron $e_i$ is then  the inner product $ \mathbf{a}_{e_i}^{\top} \mathbf{h} $.
 We consider here formalized neurons which might actually be implemented as ensembles of neurons or in other form.
 Here and in the following we assume that the matrix $A$ stores the latent representations  of all generalized entities. The context makes it clear if we refer to  the latent representations of entities, predicates, or time.
}
	\label{fig:neuronrep}
\end{figure}

The \emph{unique-representation hypothesis} assumed in this paper is that each entity or concept  $e_i$, each predicate $e_p$ and each time step $e_t$ has a  unique latent representation ---$\mathbf{a}_i$,  $\mathbf{a}_p$, respectively,  $\mathbf{a}_t$--- in form of a vector of real numbers.
The assumption is that the representations are shared between  all memory functions, and this permits information exchange and inference between the different  memories. For simplicity we assume that the dimensionalities of these latent representations are all identical $\tilde r$ such that
$\mathbf{a}_i \in \mathbb{R}^{\tilde r}$, $\mathbf{a}_p \in \mathbb{R}^{\tilde r}$, and $\mathbf{a}_t \in \mathbb{R}^{\tilde r}$.  Figure~\ref{fig:neuronrep} shows a simple  network realization.

\subsection{A Semantic Knowledge Graph Model}
\label{sec:KG}

A technical realization of a semantic memory is a knowledge graph (KG) which is a triple-oriented knowledge representation. Popular large-scale KGs are DBpedia~\cite{auer_dbpedia:_2007}, YAGO~\cite{suchanek_yago:_2007}, Freebase~\cite{bollacker_freebase:_2008}, NELL~\cite{carlson2010toward}, and the Google Knowledge Graph~\cite{singhal_introducing_2012}.

Here we consider a slight extension to    the subject-predicate-object triple form by adding the value in the form
($e_s, e_p, e_o$; \textit{Value}) where \textit{Value} is a function of $s, p, o$ and, e.g.,  can be a Boolean variable (\textit{True} or \textit{1}, \textit{False} or \textit{0})  or a real number.
Thus \textit{(Jack, likes, Mary; True)} states that Jack (the subject or head entity) likes Mary (the object or tail entity).
Note that $e_s$ and $e_o$ represent the entities for subject index $s$ and object index $o$. To simplify notation we also consider $e_p$ to be a generalized  entity associated with predicate type with index $p$. We encode attributes also as triples, mostly to simplify the discussion.

We now consider an efficient representation of a KG. With this representation, it is also possible  to generalize from known facts to new facts (inductive inference).
 First, we introduce the three-way semantic  adjacency tensor
 $\mathcal{X}$
where the  tensor element $x_{s, p, o}$  is the associated \textit{Value} of the triple ($e_s, e_p, e_o$). Here $s = 1, \ldots, S$, $p = 1, \ldots, P$, and $o = 1, \ldots, O$.
 One can also define a  companion  tensor $\underline{\Theta}$  with with the same dimensions as $\mathcal{X}$ and with entries $\theta_{s, p, o}$.  It contains the natural parameters of the model and the connection to $\mathcal{X}$  for Boolean variables
  is
 \begin{equation}\label{eq:boolean}
    P(x_{s, p, o} | \theta_{s, p, o} )    =
\textrm{sig}(\theta_{s, p, o})
 \end{equation}
where $\textrm{sig}(arg) = 1/(1+\exp(-arg))$ is the logistic function  (Bernoulli likelihood) .  If $x_{s, p, o}$ is a real number then we can use a Gaussian distribution with  $P(x_{s, p, o} | \theta_{s, p, o} ) \sim  {\mathcal{N}}(\theta_{s, p, o}, \sigma^2)$. Unless specified otherwise, we will assume a Bernoulli distribution for the rest of the paper.

{As mentioned, the key concept in embedding learning is that each entity $e$ has an $\tilde r$-dimensional latent vector representation $\mathbf{a} \in \mathbb{R}^{\tilde r}$.} In particular, the embedding approaches used for modeling KGs assume that
 \begin{equation}\label{eq:value}
 \theta^{\textit{semantic}}_{s, p, o} = f^{\textit{semantic}}(\mathbf{a}_{e_s}, \mathbf{a}_{e_p}, \mathbf{a}_{e_o}) .
 \end{equation}
Here, the function $f^{\textit{semantic}}(\cdot)$  predicts the value of the natural parameter.
In the case of a KG with a Bernoulli likelihood,   $\textrm{sig}(\theta^{\textit{semantic}}_{s, p, o})$
represents the confidence that the \textit{Value} of the  triple ($e_s, e_p, e_o$)  is true and we call the function an \textit{indicator mapping function} and we discuss examples in the next section.

Latent representation approaches have been used very successfully to model large KGs, such as the YAGO KG, the DBpedia KG and parts of the Google KG.  It has been shown experimentally that models using latent factors perform well in these high-dimensional and highly sparse domains.  Since an entity has a unique representation,  independent of its role as a subject or an object, the model permits the propagation of information across the KG.  For example if a writer was born in Munich, the model can infer that the writer is also born in Germany and probably writes in the German language~\cite{nickel_three-way_2011,nickel_factorizing_2012}. Stochastic gradient descent (SGD) is typically being used as an iterative  approach for finding both optimal latent representations and optimal parameters in $f^{\textit{semantic}} (\cdot)$~\cite{nickel2015,krompass2015type}.  For a recent review, please consult~\cite{nickel2015}.

Due to the approximation, $\textrm{sig} (\theta^{\textit{semantic}}_{\textit{Jack}, \textit{marriedTo}, e})$ might be smaller than one for the true spouse.
The approximation also permits inductive inference: We might get a large $\textrm{sig}(\theta^{\textit{semantic}}_{\textit{Jack}, \textit{marriedTo}, e})$ also for persons $e$ that are \emph{likely} to be married to \emph{Jack}
and   $\textrm{sig}(\theta^{\textit{semantic}}_{s, p, o})$  can, in general,  be interpreted as a confidence value for the triple $(e_s,e_p,e_o)$. More complex queries on semantic models involving existential quantifier are discussed in~\cite{krompas_probabilistic_2014}.

A concept memory would technically correspond to classes with  a hierarchical subclass structure. In~\cite{nickel_learning_2011,nickel2013tensor} such a structure was learned from the latent representations by hierarchical clustering. In KGs, a hierarchical structure is described by \textit{type} and \textit{subclass} relations.

Latent representations for modeling semantic memory functions have a long history in cognitive modeling, e.g., in latent semantic analysis~\cite{landauer1997solution} which is restricted to attribute-based representations.
Generalizations towards probabilistic models are probabilistic latent semantic indexing~\cite{{hofmann1999probabilistic}} and latent Dirichlet allocation~\cite{blei2003latent}.
Latent clustering and topic models~\cite{kemp_learning_2006,xu_infinite_2006,airoldi_mixed_2008} are extensions toward multi-relational domains and  use discrete latent representations.
See also~\cite{lund1995semantic,griffiths2007google,griffiths2007topics}. Spreading activation is the basis of the teachable language comprehender (TLC), which is a network model of semantic memory~\cite{collins1975spreading}.
 Associate models are the symbolic ACT-R~\cite{anderson1983architecture,anderson1997act} and  SAM~\cite{raaijmakers1981sam}.
\cite{nickel2015holographic} explores holographic embeddings with representation learning to model associative memories. An attractive feature here is that the compositional representation
has the same dimensionality as the representation of its constituents.
Connectionists  memory models are described  in~\cite{hopfield1982neural,mcclelland1985distributed,carpenter1989neural,kohonen2012self,hinton1981implementing,hinton2014parallel}.

\subsection{An Event Model for  Episodic Memory}
\label{sec:event}

Whereas a semantic  KG model reflects the state of the world, e.g, of a clinic and its patients, observations and actions describe factual knowledge about discrete events, which, in our approach,  are represented by an episodic  event tensor.
 In a clinical setting,
events might be a prescription of a medication to lower the cholesterol level,
 the decision to measure the cholesterol level
 and the measurement result  of the cholesterol level; thus events can be, e.g.,  actions,  decisions and measurements.

 The episodic event tensor is a four-way tensor
 $\mathcal{Z}$
where the  tensor element $z_{s, p, o, t}$  is the associated \textit{Value} of the quadruple ($e_s, e_p, e_o, e_t$).
  The indicator mapping function then is
 \[
\theta^{\textit{episodic}}_{s, p, o, t} = f^{\textit{episodic}}(\mathbf{a}_{e_s}, \mathbf{a}_{e_p},  \mathbf{a}_{e_o}, \mathbf{a}_{e_t})
 \]
 where
 we have added  a representation for the time of an event by introducing the generalized entity $e_t$ with latent representation $\mathbf{a}_{e_t}$. This latent representation compresses all events that happen at time $t$.

As examples, the individual can  recall ``Who did I meet last week?'' by
$e_o = \argmax_e \theta^{\textit{episodic}}_{\textit{Myself}, \textit{meet}, e, \textit{LastWeek}}$ and
``When did I meet  Jack?'' by
$e_t = \argmax_e \theta^{\textit{episodic}}_{\textit{Myself}, \textit{meet}, \textit{Jack}, e}$.

 Examples from our clinical setting would be: \textit{(Jack, orderBloodTest, Cholesterol, Week34; True)} for the fact that a cholesterol blood test was ordered in week 34 and   \textit{(Jack, hasBloodTest, Cholesterol, Week34; 160)} for the result of the blood test.
 Note that we consider an episodic event memory over different subjects,  predicates and  objects; thus episodic event memory can represent an extensive event context!


An event model can be related to the cognitive concept of an \emph{episodic memory} (Figure~\ref{fig:mem}). Episodic memory represents our memory of experiences and specific events in time in a serial form (a ``mental time travel''), from which we can reconstruct the actual events that took place at any given point in our lives~\cite{smith2013cognitive}\footnote{http://www.human-memory.net/types\_episodic.html}.
In contrast to semantic memory, it requires recollection of a prior experience~\cite{tulving1985elements}.

 For  a particular instance in time $t$, the ``slice'' of the event tensor  $\mathcal{Z}_{t}$ describes events as a, typically very sparse,  triple graph. Some of the elements of this triple graph will affect changes in the KG~\cite{Crist2015,EstebanEvent} (see also the discussion in Section~\ref{sec:shared}).
 For example the event model might record a diagnosis which then becomes a fact in the KG.
 Also the  common representations for subject, predicate, and object lead to a transfer from the event model to the semantic KG model (see also the discussion in Section~\ref{sec:shared}).

\subsection{Autobiographical Event Tensor}
\label{sec:auto}

  In some applications we want to consider the episodic information specific to an individual. For example, in a patient model, one is interested in what happened to the individual at time $t$ and not what happened to all patients at time $t$.   The  autobiographical   event tensor is simply the sub-tensor  $\mathcal{Z}_s$ concerning the events of the individual only. We then obtain  a personal time $e_{s=i, t}$ with latent representation $\mathbf{a}_{e_{s=i, t}}$. Whereas $\mathbf{a}_{e_{t}}$ is a latent representation for all events for all patients at time $t$, $\mathbf{a}_{e_{s=i, t}}$ is a latent representation for all events for patients $i$ at time $t$~\cite{Crist2015,EstebanEvent}.

The autobiographical   event tensor would correspond to the \textit{autobiographical memory}, which stores autobiographical events of an individual  on a semantic abstraction level~\cite{conway2000construction,gazzaniga2004cognitive}.
The autobiographical event tensor can be related to Baddeley's episodic buffer and, in contrast to Tulving's concept of \emph{episodic memory},   is a temporary store and is considered to be a part of working memory~\cite{baddeley2000episodic,jonides2008mind,baddeley2012working}.

\subsection{A Sensory Buffer}
\label{sec:sensor}

We assume that the sensor input consists of $Q$-channels and that at each time step $t$ a buffer is constructed of $N$ samples of the $Q$ channels. $\gamma = 0, \ldots, N$ specifies the time location within the buffer (see also Figure~\ref{fig:memHier}).
%
%
%
In contrast to the event buffer, the sensory buffer operates at a subsymbolic level. Technically it might represent measurements like temperature and pressure,  and in a cognitive model, it might represent input channels from the senses. {The sensory buffer might be related to the mini-batches in Spark Streaming where data is captured in buffers that hold seconds to minutes of the input streams~\cite{zaharia2012discretized}.}

The sensory buffer is described by  a three-way tensor
 $\mathcal{U}$
where the  tensor element $u_{q, \gamma, t}$
is the associated \textit{Value} of the triple ($e_q, e_{\gamma}, e_{t}$).
$e_q$ is a generalized entity  for the $q$-th sensory channel, $e_{\gamma}$  specifies the time location in the buffer and $e_{t}$ is a generalized entity  representing the complete buffer at time $t$.

We model
  \[
 \theta^{\textit{sensory}}_{ q,  \gamma, t} = f^{\textit{sensory}}(\mathbf{a}_{e_q},  \mathbf{a}_{e_{\gamma}}, \mathbf{a}_{e_{t}})
 \]
where $\mathbf{a}_{e_q}$  is the  latent representations for the sensor channel $e_q$ and
$\mathbf{a}_{e_\gamma}$  is the  latent representations for  $e_{\gamma}$.
Latent components corresponds to  complex time patterns (chunks) whose amplitudes are determined by the  components of $\mathbf{a}_{e_{t}}$; thus complex sensory events and sensory patterns can be modelled.

In a technical application~\cite{EstebanEvent}, the sensors measure, e.g.,  wind speed, temperature, and humidity at the location of  wind turbines and the sensory memory retains the measurements from $t-1$ to $t$.

In human cognition,
sensory memory (milliseconds to a  second) represents the  ability to retain impressions of sensory information after the original stimuli have ended~\cite{tulving1972episodic,coltheart1980iconic,gazzaniga2004cognitive}. { The transfer of sensory memory to short-term memory (e.g., the autobiographical episodic buffer) is the first step in some memory models, in particular in the modal theory of Atkinson and Shiffrin~\cite{{atkinson1968human}}. New evidence suggests that short-term memory is not the sole gateway to long-term memory~\cite{gazzaniga2004cognitive}.}
Sensory memory is thought to be located in the brain regions responsible for the corresponding sensory processing.
 Sensory memory can be the basis for sequence learning and the detection of complex time patterns.

\subsection{Comment}

The different memories and their tensor representations and models are summarized in Figure~\ref{fig:memHier}.
Under the \emph{unique-representation hypothesis} assumed in this paper, the latent representations of generalized entities are central for retrieval and prediction: the memory does not need to store all the facts and relationships about an entity.  Also, there is no need to explicitly store the semantic graph  explicitly. At any time, an approximation to the graph can be reconstructed from the latent representations. See also the discussion in Section~\ref{sec:shared}.

\subsection{Cost Functions}
\label{sec:train}

Each memory function generates a term in the cost function (see Appendix) and all terms can be considered in training to adapt all latent representations and all parameters in the various functional mappings. Note that this  is a global optimization  step involving all available data.\footnote{In human memory, one might speculate that this might be a step performed during sleep.} In general, we assumed a unique-representation for an entity, for example we assume that ${\mathbf{a}}_{e_{s}}$ is the same in the prediction model and in the semantic model.
Sometimes it makes sense to relax that assumption and only assume some form of a coupling.  Technically there are a number of possibilities: For example, the prediction model might be   trained on its own cost function,  using the latent representations from the knowledge graph as an initialization;  alternatively, one can use different weights for the different cost function terms. Some investigators propose that only some dimensions of the latent representations should be shared~\cite{alter2003generalized,acar2015data}.  \footnote{In the technical solutions~\cite{Crist2015,EstebanEvent}, we got best results by focussing on the cost function that corresponded to the problem to solve. For example in prediction tasks we optimized the latent representations and the parameters using the prediction cost function.}  \cite{larochelle2008zero,bengio2013representation,bengio2012deep} contain extensive discussions on the transfer of latent representations.
It is important to note that by considering only conditional  probability models (e.g., \textit{Value},  conditioned on  subject, predicate and object), no global normalization needs to be considered in training.

\section{Modelling the Indicator Mapping Function}
\label{sec:modelquery}

\subsection{Using General Function Approximators}

\begin{figure}
	\centering
	\includegraphics[width=\columnwidth]{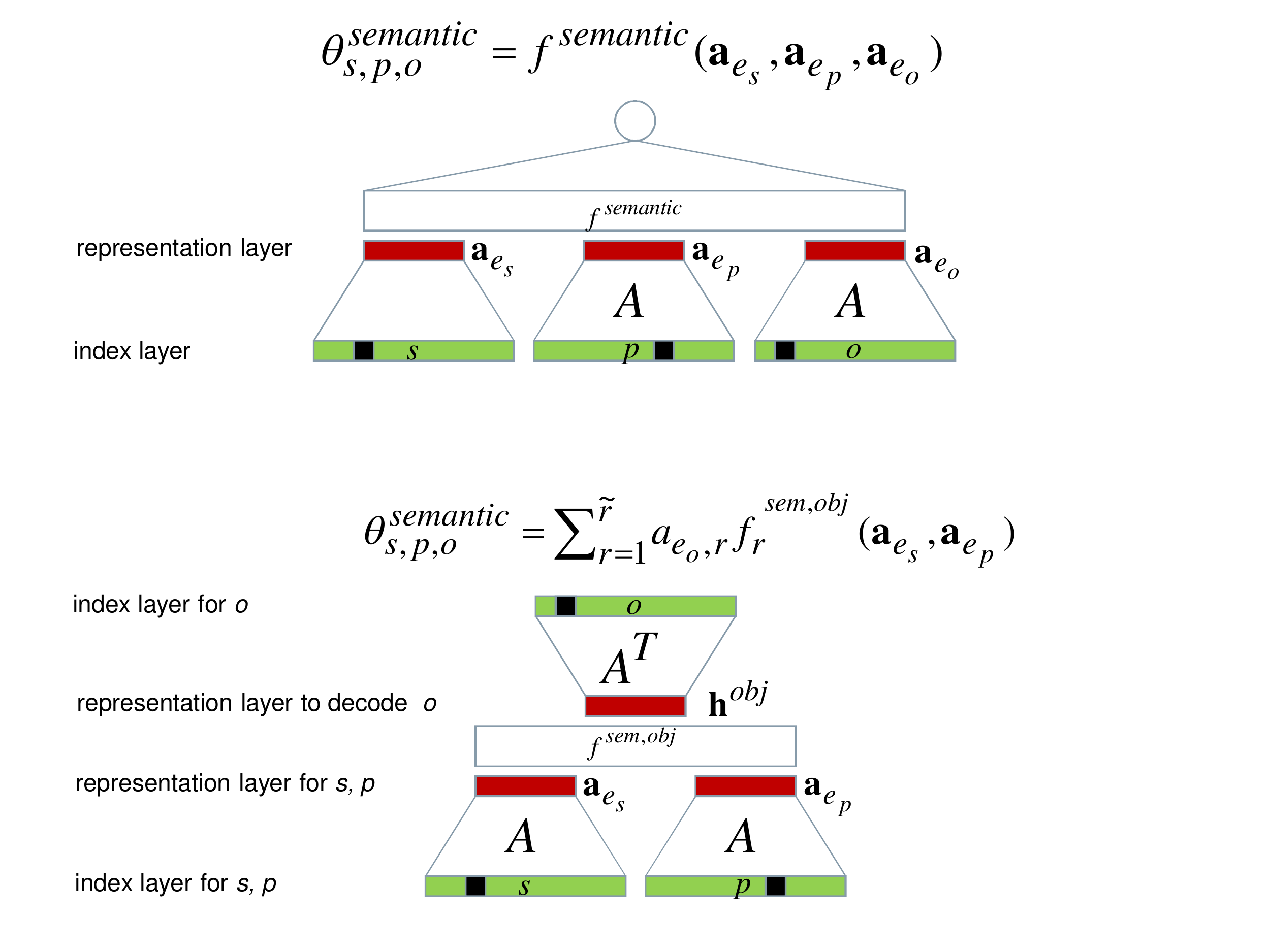}
	\caption{%
Indicator mapping function (top): The index neurons representing $e_s$, $e_p$, and $e_o$ are activated at the input  and the indicator mapping function is generated at the output. Indicator mapping  prediction using representation prediction (bottom):
With inputs   $e_s$ and  $e_p$,   a latent representation vector $\mathbf{h}^{\textit{object}}$ is calculated which activates the output index neurons encoding the objects.}
	\label{fig:ffNNx}
\end{figure}

Consider the semantic KG. Here, the indicator mapping function $f^{\textit{semantic}}(\cdot)$  can be modelled as a general function approximator, such as a feedforward ``multiway'' neural network ({NN}), where the index neurons representing $e_s$, $e_p$, and $e_o$ are activated at the input  and the response is generated at the output, as shown in the top of  Figure~\ref{fig:ffNNx}.
With this model it would be easy to query for the plausibility of a triple $(e_s, e_p, e_o; \textit{Value})$,  but other queries would be more difficult to handle.

An alternative model is shown at the bottom of  Figure~\ref{fig:ffNNx} with inputs   $e_s$ and  $e_p$  and where a function approximator predicts a latent representation vector $\mathbf{h}^{\textit{object}}$ with components
\[
h^{\textit{object}}_{r} =
 {f}_{r}^{\textit{semantic, object}} (\mathbf{a}_{e_s}, \mathbf{a}_{e_p}) \;\;\; \;\;\;  r = 1, \ldots, \tilde r .
\]
The function $f^{\textit{semantic}}(\cdot)$   is now    calculated as an inner product between the predicted latent representation and the latent representation of the objects as
 \begin{equation}\label{eq:rep-pred}
f^{\textit{semantic}}(\mathbf{a}_{e_s}, \mathbf{a}_{e_p}, \mathbf{a}_{e_o}) =
\mathbf{a}_{e_o}^\top \mathbf{f}^{\textit{semantic, object}} (\mathbf{a}_{e_s}, \mathbf{a}_{e_p})
=  \mathbf{a}_{e_o}^\top  \mathbf{h}^{\textit{object}}
.
 \end{equation}
Here, $\mathbf{f}^{\textit{semantic, object}} = ({f}_1^{\textit{semantic, object}}, \ldots, {f}_{\tilde r}^{\textit{semantic, object}})^{\top}$.

Thus the response to the query $(\textit{Jack}, \textit{likes}, ?)$ can be obtained by activating the index neurons for \textit{Jack} and \textit{likes} at the input and  by considering index neurons at the outputs with large values.
Note that with $\mathbf{f} = \mathbf{{f}}^{\textit{semantic, object}}(\cdot)$,   a function approximator produces a latent representation vector $\mathbf{h}$ and the activation of the output index neurons corresponds to the likelihood  that $e_o$ is the right answer. We call this modelling approach indicator mapping by representation prediction.

\begin{figure}
	\centering
	\includegraphics[width=\columnwidth]{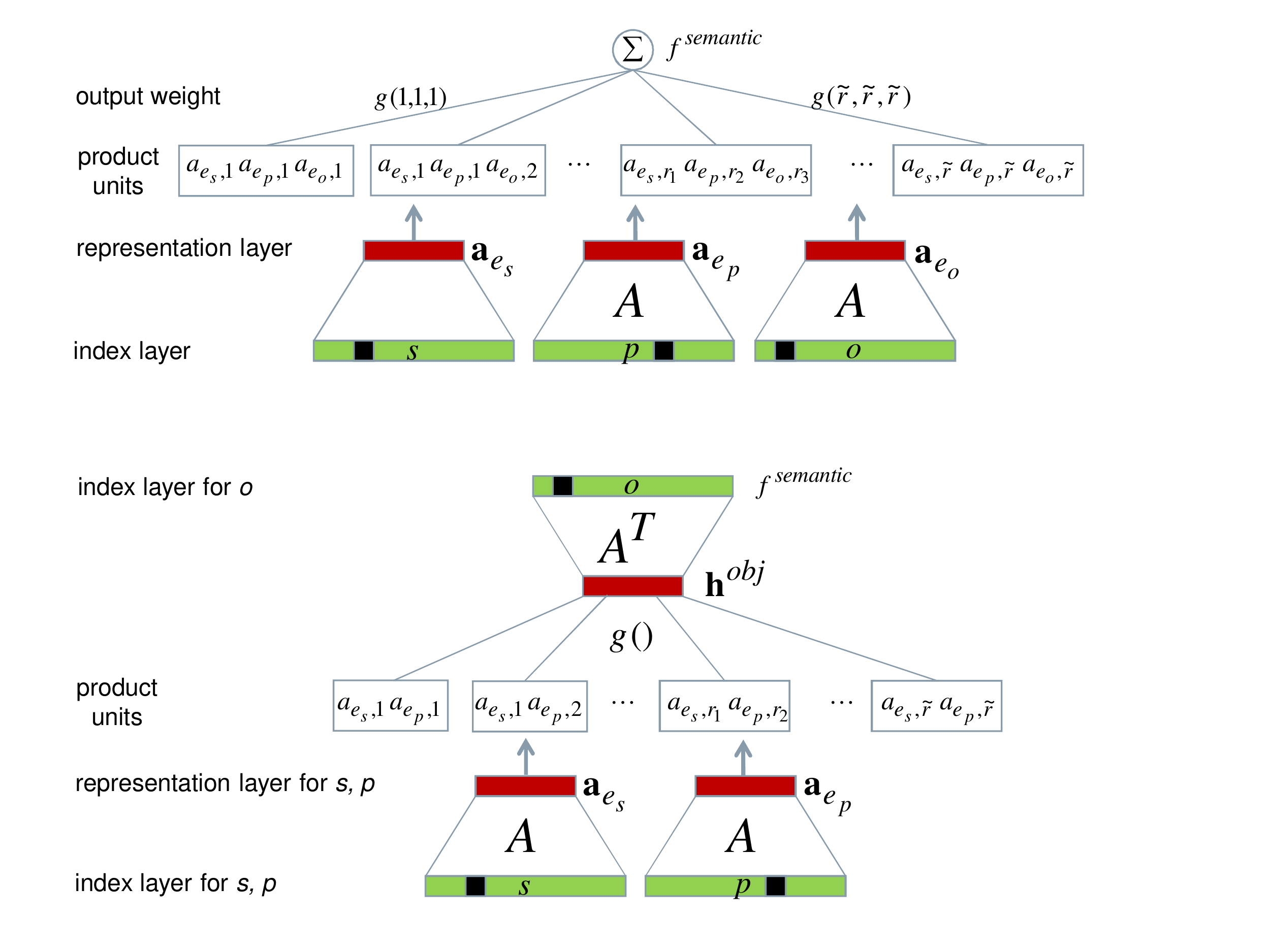}
	\caption{%
As in Figure~\ref{fig:ffNNx} but with a Tucker model. Top: An architecture with two hidden layers, Interaction between latent representations are implemented by the product nodes. Bottom: Same but drawn as a model with three hidden layers. The $g(\cdot)$-layer fully connects the outputs of the product layer with the object  representation layer. Since the Tucker  model is symmetrical with respect to  the generalized entities, in the following we draw all representations below the $g(\cdot)$-layer.  }
	\label{fig:ffNNxtensor}
\end{figure}

\begin{figure}
	\centering
	\includegraphics[width=\columnwidth]{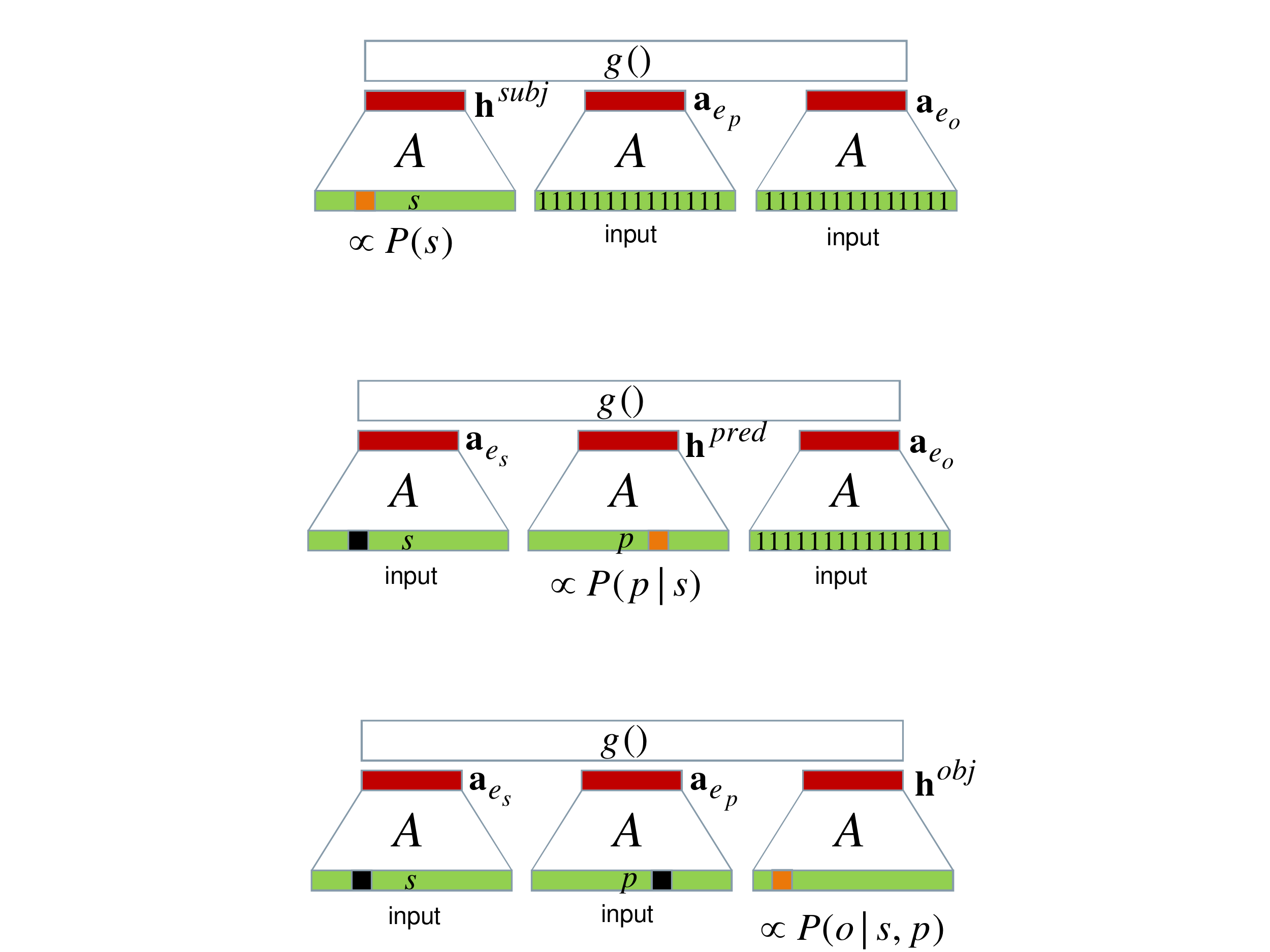}
	\caption{%
For nonnegative tensor models, marginals and conditionals can easily be calculated and independent samples from the model can be calculated. The figure shows the situation for a Tucker model. In the top model, we apply vectors of ones to the predicate and object representation, which leads to a marginalization of those variables. The subject representation acts as output and we can sample a subject. In the center, we only integrate out the object and use the subject index as input. At the predicate output we can sample a predicate. Finally, in the bottom, we use subject and predicate samples as inputs and produce a sample for an object. Naturally, when subject and predicate are given, we only need to use the model at the bottom.
}
	\label{fig:ffNNy}
\end{figure}

\subsection{Tensor Decompositions}

Tensor decompositions have also shown excellent performance in modelling KGs~\cite{nickel2015}. In tensor decompositions, the indicator mapping function  $f^{\textit{semantic}}(\cdot)$ is implemented as a multilinear model.

    Of particular interest are the  PARAFAC model (canonical decomposition) with
\[
 f^{\textit{semantic}}(\mathbf{a}_{e_s}, \mathbf{a}_{e_p}, \mathbf{a}_{e_o}) = \sum_{r=1}^{\tilde r}  {a}_{e_s, r} \;  {a}_{e_p, r} \;  {a}_{e_o, r}
\]
and the Tucker model with
\[
 f^{\textit{semantic}}(\mathbf{a}_{e_s}, \mathbf{a}_{e_p}, \mathbf{a}_{e_o}) =
 \sum_{r_1=1}^{\tilde r} \sum_{r_2=1}^{\tilde r} \sum_{r_3 =1}^{\tilde r}
 {a}_{e_s, r_1}   \;   {a}_{e_p, r_2}   \;  {a}_{e_o, r_3}  \;
 g(r_1, r_2, r_3)  .
\]
Here, $g(r_1, r_2, r_3) \in \mathbb{R}$ are elements of the core tensor $\mathcal{G} \in R^{\tilde{r} \times \tilde{r} \times \tilde{r}}$.  Finally, the RESCAL model~\cite{nickel_three-way_2011} is a Tucker2 model with
\[
 f^{\textit{semantic}}(\mathbf{a}_{e_s}, \mathbf{a}_{e_p}, \mathbf{a}_{e_o}) =
\sum_{r_1=1}^{\tilde r} \sum_{r_2=1}^{\tilde r}
  {a}_{e_s, r_1}  \;  {a}_{e_o, r_2}  \;  g(r_1, r_2, e_p)
\]
with core tensor $\mathcal{G} \in R^{\tilde{r} \times \tilde{r} \times {P}}$. In all these models, we use the constraint that a generalized  entity has a unique latent representation.

An attractive feature of tensor decompositions is that,  due to their multilinearity,   representation prediction models can  easily be constructed: For the PARAFAC model,
$h^{\textrm{object}}_{r}  =   {a}_{e_s, r} \;  {a}_{e_p, r}$,
 for Tucker,  $h^{\textrm{object}}_{r}  =    \sum_{r_1=1}^{\tilde r} \sum_{r_2=1}^{\tilde r}
 {a}_{e_s, r_1}     {a}_{e_p, r_2} \;
 g(r_1, r_2, r)$
 and for RESCAL
$
h^{\textrm{object}}_{r}   =
\sum_{r_1=1}^{\tilde r}
  {a}_{e_s, r_1}  \;   \;  g(r_1, r, e_p)
$.  The architectures for the Tucker model are drawn in Figure~\ref{fig:ffNNxtensor}.

\section{Querying Memories}
\label{sec:qa}

\subsection{Function Approximator  Models}

In many application one is interested in retrieving triples with a high likelihood, conditioned on some information, thus we are essentially faced with an optimization problem. To answer a query of the form $(\textit{Jack}, \textit{likes}, ?)$ we need to solve
\[
\arg \max_{\mathbf{a}_{e_o}} f^{\textit{semantic}}(\textit{Jack}, \textit{likes}, \mathbf{a}_{e_o})  .
\]
Of course one is often interested in a set of likely answers.

We suggest to address querying via a simulated annealing approach. We define an energy function
$E(s, p, o) = - f^{\textit{semantic}}(\mathbf{a}_{e_s}, \mathbf{a}_{e_p}, \mathbf{a}_{e_o})$  and define a Boltzmann distribution as
\[
P(s, p, o) = \frac{1}{Z(\beta)} \exp  \beta  f^{\textit{semantic}}(\mathbf{a}_{e_s}, \mathbf{a}_{e_p}, \mathbf{a}_{e_o}) .
\]
Here $Z(\beta)$ is the partition function that normalizes the distribution and $\beta \ge 0$ is an inverse temperature.
Note that we now have generated a probability distribution where subject, predicate, and object are the random variables!\footnote{Previously, only the \textit{Value } conditioned on subject, predicate, and object was random.}

Now to answer the query,  $(\textit{Jack}, \textit{likes}, ?)$, we sample from
\[
P(o | s,  p) =
\frac{1}{Z(s, p, \beta)}
 \exp  \beta  f^{\textit{semantic}}(\mathbf{a}_{e_s}, \mathbf{a}_{e_p}, \mathbf{a}_{e_o})
\]
with $s = \textit{Jack}$ and $p = \textit{likes}$.
The artificial inverse temperature $\beta \ge 0$ can determine if we are interested in just sampling the most likely response (large $\beta$) or are also interested in responses with a smaller probability (small $\beta$). Similarly, we can derive models for $P(s | p, o)$ and  $P(p | s, o)$.\footnote{In the Appendix in Subsection~\ref{sec:sfa}
 (Figure~\ref{fig:funcspo}) we describe how samples from $P(s)$, $P(p|s)$, and $P(o|s, p)$ can be obtained.}

\subsection{Tensor Models}

By enforcing nonnegativity of the factors and the core tensor entries,  we can define a  probabilistic model for a Tucker model with
$E(s, p, o) = - \log f^{\textit{semantic}}(\mathbf{a}_{e_s}, \mathbf{a}_{e_p}, \mathbf{a}_{e_o})$ as
\begin{equation}\label{eq:tenen}
  P(s, p, o) \propto
\left(
 \sum_{r_1=1}^{\tilde r} \sum_{r_2=1}^{\tilde r} \sum_{r_3 =1}^{\tilde r}
 {a}_{e_s, r_1}   \;   {a}_{e_p, r_2}   \;  {a}_{e_o, r_3}  \;
 g(r_1, r_2, r_3)
  \right)^{\beta}  .
\end{equation}

 An attractive feature of tensor models is that  marginals and conditionals  can easily be obtained. Here, we  look at the Tucker model.  For $P(o|s,p)$ we we can use the Equation~\ref{eq:tenen} with  appropriate normalization.
 For $P(p|s)$  we use the same equation  where we replace
 ${\mathbf{a}}_{e_o}$ with $\mathbf{\bar{{a}}}^{\textit{object}} = \sum_o {\mathbf{a}}_{e_o}$.
 For $P(s)$  we use the same equation again where we replace in addition ${\mathbf{a}}_{e_p}$ with $\mathbf{\bar{{a}}}^{\textit{predicate}} = \sum_p {\mathbf{a}}_{e_p}$. As shown  in the architecture in Figure~\ref{fig:ffNNy}, these operations can easily be implemented. Marginalization means that the index neurons are all active, indicated by the vector of ones in the figure.\footnote{Note that to derive the equations for marginalization and conditioning we work with $\beta=1$; $\beta \ne 1$ is relevant during sampling. }

We can use these models to generate samples from the distribution by first generating a sample for $s$ from $P(s)$, then a sample from $p$ from $P(p|s)$, and finally a sample from $o$ using $P(o|s, p)$. {By repeating this process we can obtain independent samples from $P(s, p, o)$!}

Note that there is a certain equivalence between tensor models and sum-product networks, where similar operations for marginals and conditionals can be defined~\cite{poon2011sum}.

We can generalize the approach to all memory functions by defining suitable energy functions. We want to emphasize that we use the probability distributions only for query-answering and not for learning!

\begin{figure}
	\centering
	\includegraphics[width=\columnwidth]{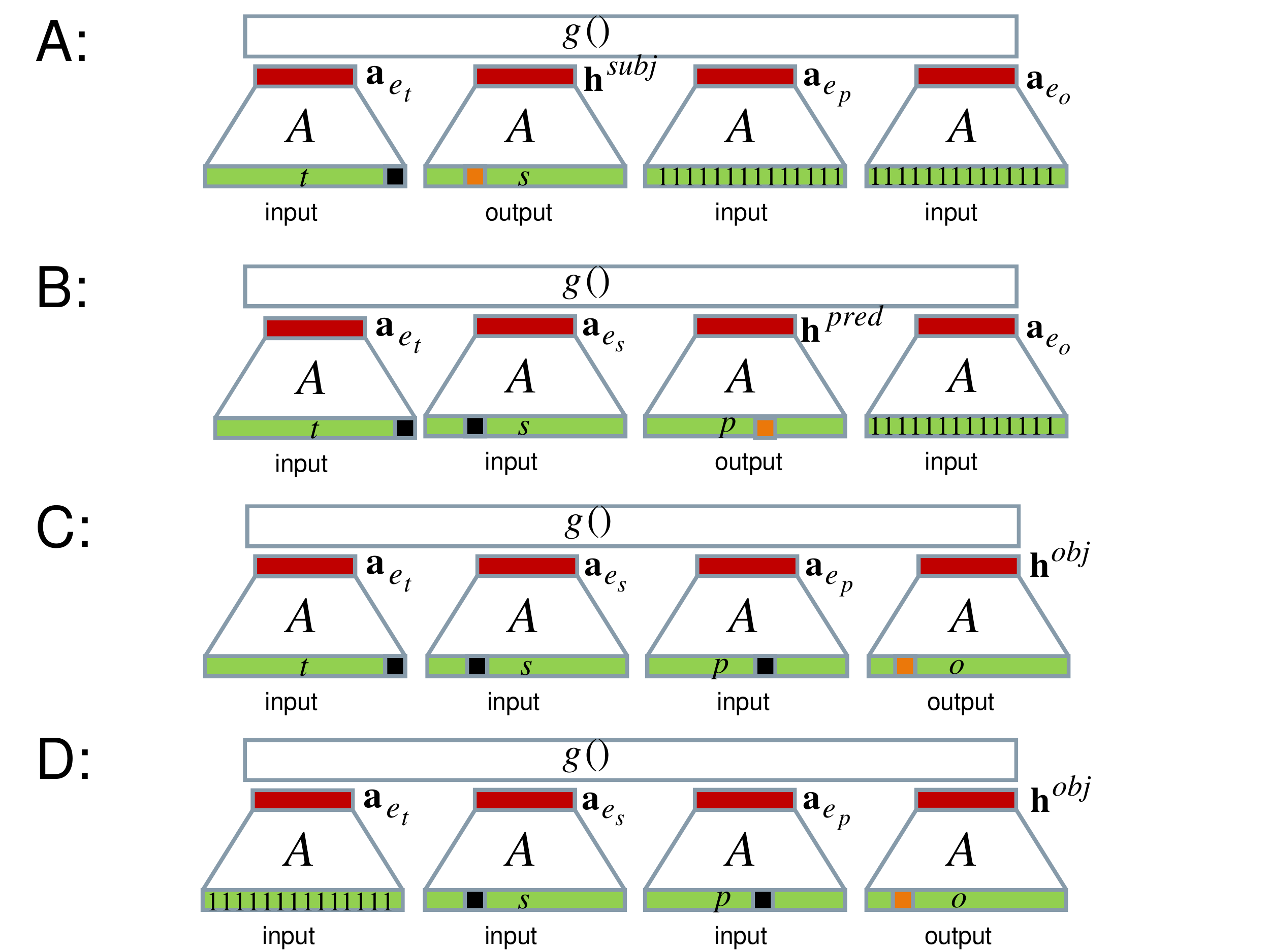}
	\caption{%
 The semantic decoding using a 4-dimensional Tucker tensor model.
  A: $\mathbf{a}_{e_t}$ is generated by the mapping of the sensory buffer by $ \mathbf{f}^M(\cdot)$. To sample a subject $s$ given time $t$,
 predicate $p$ and object  $o$ are marginalized.
  B:  Here,   $o$ is marginalized and one samples a predicate $p$, given $t, s$.
  C: Sampling of an object $o$, given $t, s, p$.
  D: By integrating out the time dimension, we obtain a memory, which is a particular  semantic memory.
  For marginalization, one can either input a vector of ones (as shown) or one learns a mean representation vector $\mathbf{\bar a}$.
}
	\label{fig:SensorChain}
\end{figure}

\section{From Sensory Memory  to Semantic Decoding}
\label{sec:senssem}

\begin{figure}
	\centering
	\includegraphics[width=0.8\columnwidth]{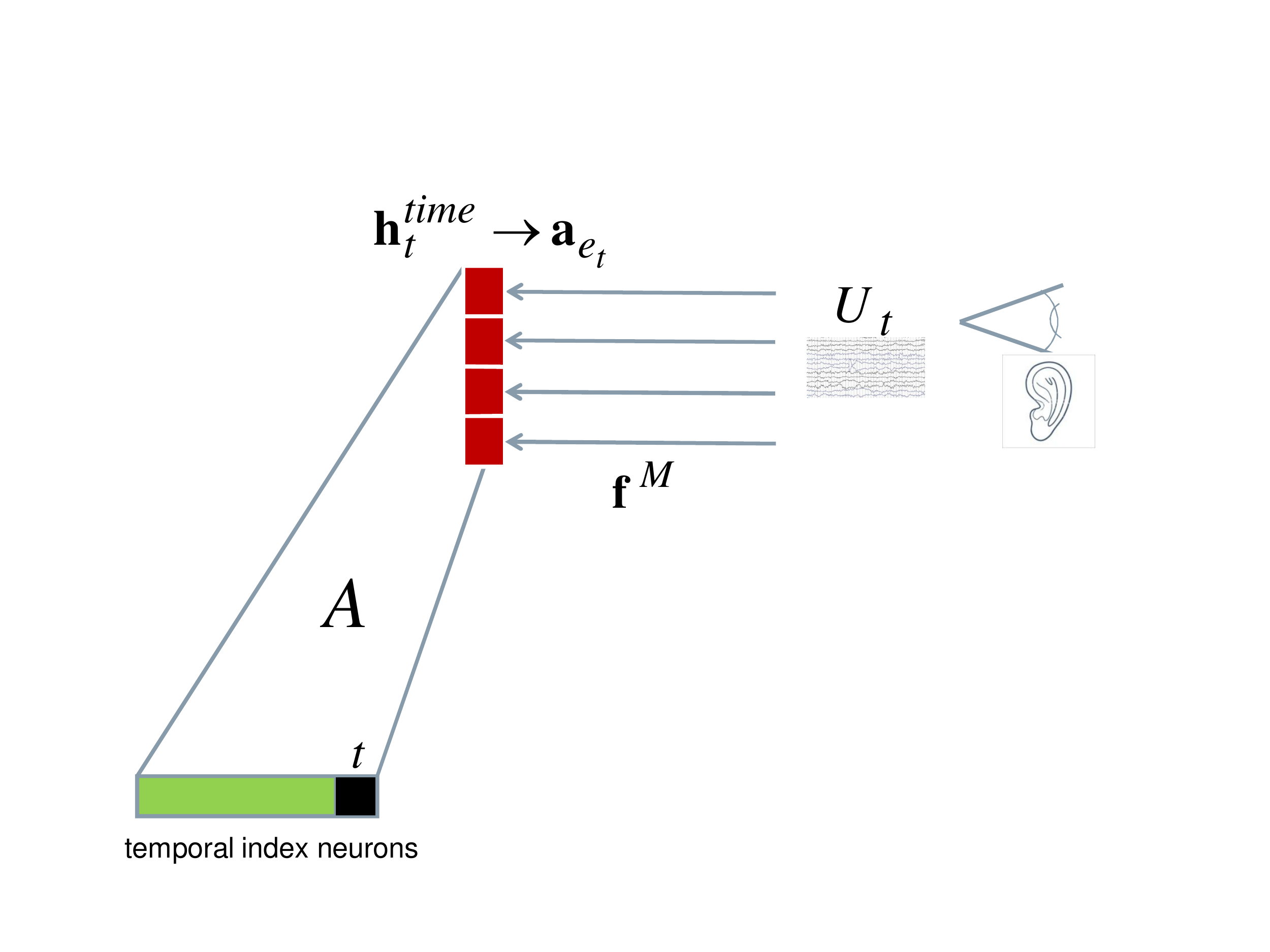}
	\caption{%
		{Mapping of $U_t = u_{:, :, t}$, i.e.,   the sensory input at time $t$,
 to the latent representation for time $\mathbf{h}_t^{\textit{time}}$
  by the function $ \mathbf{f}^M( \textrm{vec}(u_{:, :, t}) )$. If the sensory input is significant, e.g. novel, unexpected or attached with emotions, then the time
 index neuron $e_t$ is generated which stores $\mathbf{h}_t^{\textit{time}}$  as the latent representation   $\mathbf{a}_{e_t}$. These then  eventually become part of the long-term episodic memory. As indicated, $\mathbf{f}^M(\cdot)$ might consist of several sub-functions which extract different latent features.}
	}
	\label{NN-s2s}
\end{figure}

We now consider the situation that a new sensor input becomes available for time $t$. With all other latent representations and functional mappings fixed, the challenge is to calculate a new latent representation  $\mathbf{h}_t^{\textit{time}}$.
Since for a new sensory input at time $t$,  the only available information is
the sensory buffer $u_{:, :, t}$  there is a clear information propagation from sensory input to the episodic memory.
We assume a nonlinear map of the form
\begin{equation}\label{eq:sensMod}
  \mathbf{h}_t^{\textit{time}} =   \mathbf{f}^M( \textrm{vec}(u_{:, :, t}) )
\end{equation}
where $\mathbf{f}^M(\cdot)$  is a function to be learned~\cite{yinchong2016} (see Figure~\ref{NN-s2s})  and where
$\textrm{vec}(u_{:, :, t})$ are vectorized representations from the portion of the sensory tensor associated with the individual at time  $t$. Depending on the application, $\mathbf{f}^M( \cdot)$ can be a simple linear map, or it can be a second to last layer in a deep neural network as in the face recognition application \emph{DeepFace}~\cite{taigman2014deepface,ngiam2011multimodal}. In general, we assume that $\mathbf{f}^M( \cdot)$ is realized by a set of functions, where each function focusses on different aspects of the sensory inputs (Figure~\ref{NN-s2s}).  For example, if the sensory input is an image, one function might analyse color, another ones shape and a third one texture.

 One can think of $\mathbf{h}_t^{\textit{time}}$ as the latent representation of a query;  the decoding in the semantic decoder then corresponds to the answer to the query.

Assuming that a Tucker model is used for decoding, the conditional probability  becomes\footnote{To ensure nonnegativity one might want to model $\mathbf{h}_t^{\textit{time}} = \exp  \mathbf{f}^M( \textrm{vec}(u_{:, :, t}) )$.}\footnote{This equation describes a special form of a  conditional random field. With proper local normalization it becomes a conditional  multinomial probabilistic   mixture model.}
\begin{equation}\label{eq:tenenU}
  P(s, p, o | \textrm{vec}(u_{:, :, t})) \propto
\left(
 \sum_{r_1=1}^{\tilde r} \sum_{r_2=1}^{\tilde r} \sum_{r_3 =1}^{\tilde r} \sum_{r_4 =1}^{\tilde r}
    {a}_{e_s, r_1}   \;   {a}_{e_p, r_2}   \;  {a}_{e_o, r_3}  \; {h}_{t, r_4} \;
 g(r_1, r_2, r_3, r_4)
  \right)^{\beta}  .
\end{equation}
A sampling approach for decoding with a Tucker model is shown in Figure~\ref{fig:SensorChain}.

{For general function approximators one needs to train separate models for the different conditional and marginal  probabilities, as discussed in Subsection~\ref{sec:sfa} (Figure~\ref{fig:SensorCh}).}


Note that in the decoding step we transfer  information from a subsymbolic  sensory representation to a symbolic semantic representation.

{Also note that,  in pure perception, no learning of any kind needs to be involved. Only when   the sensory input is significant, e.g. novel, unexpected or attached with emotions, then the time
 index neuron $e_t$ is generated which stores $\mathbf{h}_t^{\textit{time}}$  as its latent representation   $\mathbf{a}_{e_t}$. By this operation an episode or event is generated. The time index neuron and its latent representation
are eventually transferred to  long-term episodic memory (Figure~\ref{NN-s2s}).}


\begin{figure}
	\centering
	\includegraphics[width=0.8\columnwidth]{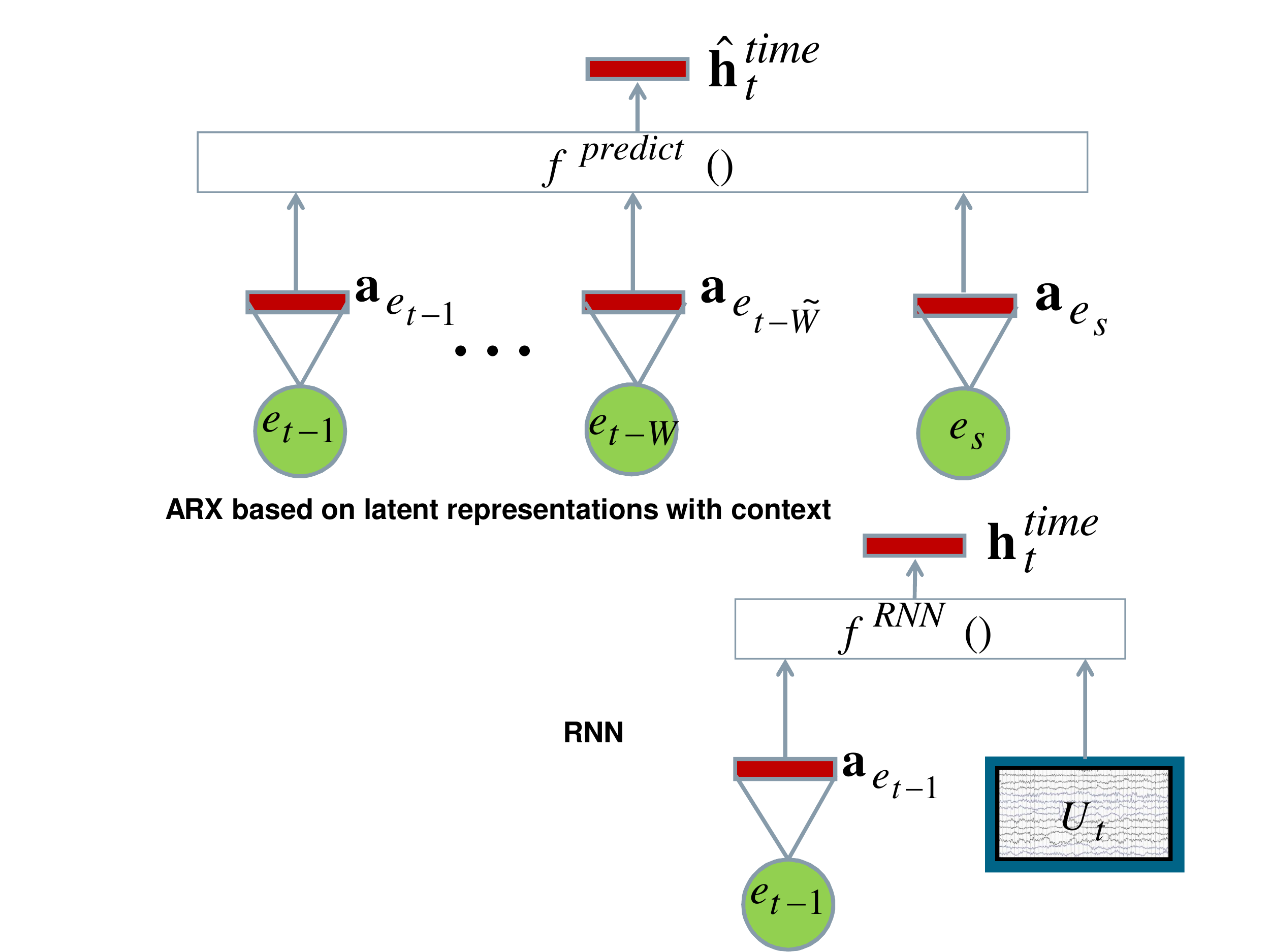}
	\caption{%
		Two different prediction models.  In the ARX model on top, we assume that $\mathbf{\hat h}$ is a deterministic function of  the sensory input $u_{:, :, t}$ and \textit{not} of past latent states. Time dependencies on $u_{:, :, t}$ are reflected in time dependencies on the latent states $\mathbf{\hat h}$ and thus future values of the latent states can be predicted using past latent states. The RNN on the bottom corresponds to the dependencies typically used in recurrent neural networks and state space models. Here past latent states  causally influence future latent states.
	}
	\label{NN-arch}
\end{figure}

\section{Predictions with Memory Embeddings and   Working Memory}
\label{sec:predict}

In this section we focus on working memory, which orchestrates the different memory functions, e.g.  for prediction and decision making. In a way working memory represents the intelligence on top of the memory functions and links to complex decision making and  consciousness have been made. Here we will focus on the restricted but  important task of prediction.
For example,  in a clinical setting, it is important to know what should be done next (e.g., prediction of a medical procedure) or what event will happen next  (e.g., prediction of a medical diagnosis).

We propose that prediction should be happen at the level of the latent representation for time, i.e.,
$\mathbf{\hat h}$, which is  the output of the sensory map,  and we consider two cases.

\subsection{ARX Model for Predicting Latent Representations of Time}

Here we assume that $\mathbf{\hat h}$ is a deterministic function of the sensory input via Equation~\ref{eq:sensMod} but not of past  time  latent representations. There might be time dependencies in the sensory input;  due to high dimensionality of the input, it is easier to model the dependencies between the latent representations instead,  as
\[
\mathbf{\hat h}_t^{\textit{time}}
=  \mathbf{f}^{\textit{predict}}(\mathbf{a}_{e_{t-1}} , \mathbf{a}_{e_{t-2}} , \ldots, \mathbf{a}_{e_{t-W}} , \mathbf{a}_{e_{\textit{indiviual}}}) .
\]
But note that this model is only used for prediction $\mathbf{h}_t^{\textit{time}}$ and as soon as  the sensory input is available, it overrides the prediction with Equation~\ref{eq:sensMod}! The model is  also suitable  for novelty detection: if $\mathbf{\hat h}$  is different from $\mathbf{h}_t^{\textit{time}}$, then the sensory scene might be novel.

Note that we also include the latent representation of the individual $\mathbf{a}_{e_{\textit{indiviual}}}$ which can be interpreted as a representation of the state of the individual.

The model can be interpreted as an autoregressive model on the latent representations with external inputs, ARX (Figure~\ref{NN-arch}, top). The parameter $\tilde W$ is the size of the time window and might be related to the capacity of short-term memory, i.e., the number of items the  working memory can consider in  decision making.

\subsection{Recurrent Model}

Here we extend the model is Equation~\ref{eq:sensMod2} to include past information of the latent representation as
\begin{equation}\label{eq:sensMod2}
  \mathbf{h}_t^{\textit{time}}  =   \mathbf{f}^{RNN}( \textrm{vec}(u_{:, :, t}), \mathbf{a}_{e_{t-1}}, \mathbf{a}_{e_{\textit{indiviual}}}) .
\end{equation}
Note that this is the structure of a recurrent neural network and the assumption is that the latent state depends on both sensory input and the previous  latent state. The architecture is shown in Figure~\ref{NN-arch}, bottom.

Both models are reasonable for different purposes and make different assumptions. {In fact, both models might play a role in human cognition. }

Alternatively one might use  networks with additional memory buffers and attention mechanisms~\cite{hochreiter1997long,weston2014memory,graves2014neural,kumar2015ask,goodfellow2015deep}.

\section{Hypotheses on Human Memory}
\label{sec:shared}

\begin{figure}
	\centering
	\includegraphics[width=\columnwidth]{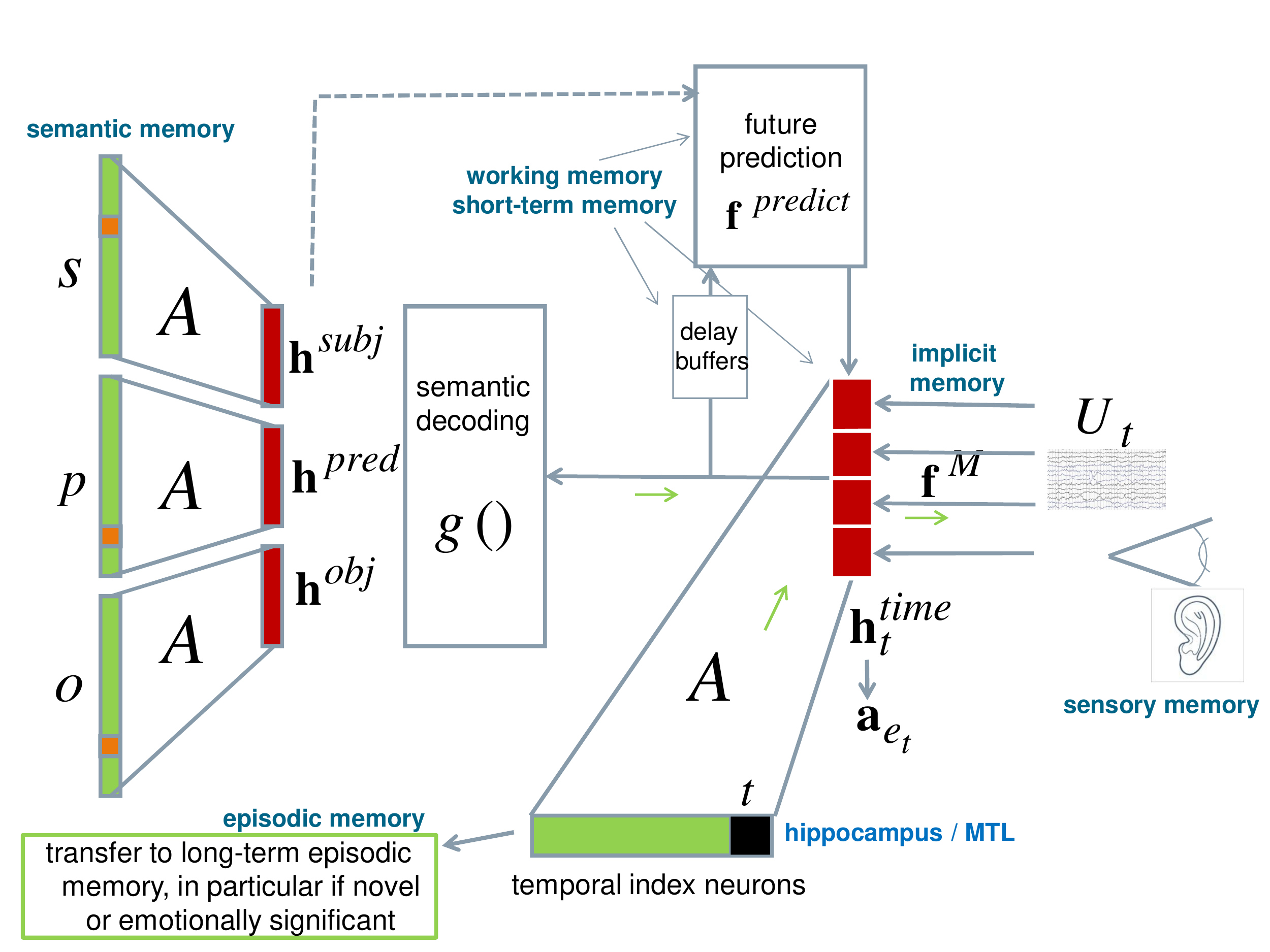}
	\caption{%
		A model for human memory. First consider the semantic decoding step. $U_t = u_{:, :, t}$ is the sensory buffer at time $t$. $ \mathbf{f}^M(\cdot)$ maps the sensory buffer to $\mathbf{h}_t^{\textit{time}}$.
 As discussed before,  this function might be realized by a set of modules where each module focusses on certain aspects of the sensory input.
{ If the memory is novel or emotionally significant, a new episodic memory is formed by the generation of time index neuron $e_t$; its latent representation is then stored as weight pattern $\mathbf{a}_{e_t} = \mathbf{h}_t^{\textit{time}}$.    The index neuron and the representations can eventually  become part of long-term episodic  memory.}
 The  semantic decoding module  (here: Tucker tensor model) then produces highly probable $(s, p, o)$-triples, given $\mathbf{h}_t^{\textit{time}}$ and as described in Figure~\ref{fig:SensorChain}. {Semantic decoding also also performed when no episodic memory is formed. }
  A form of a semantic memory can be achieved by marginalizing  time.
It is even possible to operate the model in reverse: If we consider $s$ to be the input, let's say \textit{Mary}, marginalize out $p$ and $o$ and consider $\mathbf{h}_t^{\textit{time}}$ as the output, then we can recall when we met \textit{Mary}, by exciting the time index neuron, and we can even recall how \textit{Mary} looked like and sounded like  by operating $\mathbf{f}^M(\cdot)$ in reverse. The reverse direction is indicated by the small green arrows in the figure.  Similarly, a time index neuron on the bottom can excite $\mathbf{h}_t^{\textit{time}}$, and a past scene is both semantically analysed and a sensory impression can be recalled.
$f^{predict} (\cdot)$  predicts future $\mathbf{h}_t^{\textit{time}}$ and can be used for the prediction of events and decisions and for novelty detection (Figure~\ref{fig:SensorChain}, bottom). As before, the predicted $\mathbf{h}_t^{\textit{time}}$ can be semantically decoded and can lead to mental imagery, permitting an analysis of expected events and sensory inputs.  For learning,  model parameters are adapted to facilitate the semantic decoding. If needed, representations for new generalized entities are introduced. The blue labels, which refer to human memories,  naturally are more or less speculative. {Note, that in the figure we draw different index neurons for entities in their roles as subject and object. In a way this is an artefact of the visualization of the sampling process. We maintain the hypothesis that an entity has a unique index neuron and a unique latent representation.}  }
	\label{fig:NN-bio}
\end{figure}

This section speculates about the  relevance of the presented models to  human memory functions. In particular we present several concrete hypotheses.
Figure~\ref{fig:NN-bio} shows the overall model and explains the flow of sensory input to long-term memory and semantic decoding.

\subsection{Triple Hypothesis}

A main assumption of course is that semantic memory is described by triples,  and that episodic memory is described by triples in time, i.e., quadruples.  In a way this is the perspective from which this paper has been written.  Arguments for this representation are that higher-order relations can always be reduced to triples and that triple representations have large practical significance and have been used in
 large-scale KGs.

\subsection{Unique-representation Hypothesis for Entities and Predicates}

The \textit{unique-representation hypothesis} states that
each generalized entity $e$  is represented by an index neuron and a
 unique (rather high-dimensional) latent representation $\mathbf{a}_e$  that is stored as weight patterns connecting the index neurons with neurons in the representation layer (see Section~\ref{sec:memtens} and shown in Figure~\ref{fig:neuronrep}). Note that the weight vectors might be very sparse and in some models nonnegative.
  They are  the basis for episodic memory and semantic memory.
 The latent representations integrate all that is known about a generalized  entity and can be instrumented for prediction and decision support in working memory. Among other advantages, a common representation would explain why  background information about an entity is seemingly effortlessly integrated into sensor scene  understanding and decision support by humans, at least for entities familiar to the individual.

Researchers have  reported on a remarkable subset of medial temporal lobe (MTL) neurons that are selectively activated by strikingly
different pictures of given individuals, landmarks or objects and in some cases even by letter strings with their names~\cite{quiroga2012concept,quiroga2005invariant}.
For example, neurons have been shown to selectively respond  to  famous actors like ``Halle Berry''.
Thus a local encoding of index neurons  seems biologically plausible.


As stated before, we do not insist that index neurons representing single entities exist as such in the brain, rather that there is  a level of abstraction, which is equivalent to an index neuron, e.g., an ensemble of neurons.

Our hypothesis supports both locality and globality of encoding~\cite{mcclelland1985distributed,edelman1992bringing}, since index neurons are local representations of generalized entities, whereas  the representation layers  would be high-dimensional and non-local.

 Figure~\ref{fig:NN-bio}  shows index layers and representation layers for entities and relation types on the left.
{Note, that in the figure we draw different index neurons for entities in their roles as subject and object. In a way this is an artefact of the visualization of the sampling process.  We maintain the hypothesis that an entity has a unique index neuron and a unique latent representation.}

An interesting question is if the latent dimensions have a sensible and maybe useful interpretation, which the brain might exploit!

Often  neurons with similar receptive fields are clustered together in sensory cortices and form a topographic map~\cite{gluck2013learning}.  Topological maps might also be the organizational form of neurons representing entities. Thus,  entities with similar latent representations might be topographically close. { A detailed  atlas of semantic categories  has been established in extensive fMRI studies showing the  involvement of
the  lateral temporal cortex (LTC), the ventral temporal cortex (VTC),
the lateral parietal cortex (LPC),  the medial parietal cortex (MPC),
the medial prefrontal cortex, the superior prefrontal cortex (SPFC) and the inferior prefrontal cortex
(IPFC)~\cite{Huth2016}.}

 Although the established assumption is that no new neurons are generated in the adult cortex,  topographic maps might change, e.g., due to injury, and exhibit considerably plasticity.  Consequently, one might speculate that index neurons for novel entities not yet represented in the cortex need to be integrated in the existing topographic organization.   This would not be a contradiction to our model, since,  although we require some representation for index neurons, it is irrelevant which individual neurons represent which entities. {Index and representation neurons for new entities might be allocated in the hippocampus, although,  and their function later be transferred  to the cortex.}

\subsection{Representation of Concepts}

{So far our discussion focussed on generalized entities and their latent representations and similarity between entities was expressed by the similarity in their latent representations.  In contrast, machine learning is typically concerned with the assignments of entities to concepts.  Concepts bring a certain order: for example one can imply certain properties by knowing that \textit{Cloe} is a cat.  Concept learning is not the main focus of this paper and we only want to describe one simple realization. Consider that we treat a concept simply as another entity with its own latent representation, as, e.g., in~\cite{nickel_factorizing_2012}.  We can  introduce the relation type \textit{type}, which links entities with their concepts. The inductive inference during model learning can then materialize that \textit{Cloe} is also a mammal and a living being and that, by default,  it has typical cat-attributes.  }

\subsection{Spatial Representations}

In our proposed model, we can treat locations just as any other entity. An example would be
$(\textit{Mary}, \textit{observedIn}, \textit{TownHall}, \textit{LastFriday})$. To model that the individual her- or himself was at the Townhall last Friday, a triple would be sufficient such as $(\textit{meLocation}, \textit{TownHall}, \textit{LastFriday})$ and an individual's spatial decoding might be  done  by a dedicated circuitry separate  from semantic decoding.

\subsection{Sensory Input is Transformed into a Latent Representation for  Time}

In our model we assume that  each sensory impression is decoded into a  time latent representation  $\mathbf{h}_t^{\textit{time}} = \mathbf{a}_{e_t}$ by $M$-map $\mathbf{f}^M( \cdot)$, which actually might be implemented as a set of modules, responsible for different aspects of the sensory input.

Thus, $\mathbf{h}_t^{\textit{time}}$ is a representation shared between the sensory buffer and the episodic memory and might play a role in  the phonological loop and the visuospatial sketchpad.
$\mathbf{f}^M( \cdot)$ is the most challenging component in the system.\footnote{A simple special case is when $u_{:,:, t}$ already is on a semantic level. This is the case  in the medical application described in~\cite{EstebanEvent,Crist2015} where $u_{:,:, t}$  describes procedures and  diagnosis and one can think of $\mathbf{f}^M( \cdot)$  as being an encoder system and $\mathbf{f}^{\textit{episodic}}( \cdot)$ as being a decoder and the  complex as being an autoencoder~\cite{bourlard1988auto,hinton1994autoencoders}.}
 The training of $\mathbf{f}^M( \cdot)$ to refine its operation would correspond to  perceptual learning in cognition.
 In the brain,  $\mathbf{f}^M( \cdot)$ would likely be implemented by the different sensor pathways, e.g., the visual pathway and the  auditory pathway and could contain internal feedback loops.
 Note that we would assume that the connection between the sensory representation and the time-representation  is to some degree bi-directional, thus the time representation also feeds back to sensory impressions.


\subsection{New Representations are  formed  in the Hippocampus and are then Transferred  to Long-Term Episodic and Semantic Memories}
\label{sec:lare}

{If sensory impressions are significant,  a time  index neuron $e_t$
 is formed and  sensory information is quickly implemented as a weight pattern $\mathbf{a}_{e_t} = \mathbf{h}_t^{\textit{time}}$,  as shown in Figures~\ref{NN-s2s} and~\ref{fig:NN-bio}. The time index neurons  might be ordered sequentially, so the brain maintains a notion of temporal closeness and temporal order.
Index neurons for time, i.e.,  $e_t$,  might be formed in the hippocampal region of the brain.   Evidence for time cells have  recently been found~\cite{eichenbaum2012towards,eichenbaum2014time,kitamura2015entorhinal,kitamura2015entorhinal2}. {It has been observed that the hippocampus becomes activated when the temporal order of events is being processed~\cite{lehn2009specific,rolls2010noisy,rolls2010computational}.}
Our model is in accordance with the concept that perceived sensations are decoded in the various sensory areas of the cortex, and then combined in the brain’s hippocampus into one single experience.}

According to our proposed model, the hippocampus would need to assign new time neurons during lifetime.  In fact, it has been observed that the adult macaque monkey forms a few thousand new neurons daily~\cite{gluck2013learning,gould1999neurogenesis}, possibly to encode new information~\cite{becker2005computational}. {Neurogenesis has been established in  the dentate gyrus (part of the hippocampal formation) which is thought to contribute to the formation of new episodic memories.
}

The hippocampus might be the place where new index neurons and representations are generated in general, i.e., also for new places and entities. Certainly,  the hippocampus is involved in forming new spatial representations. There are  multiple, functionally specialized, cell types of the hippocampal-entorhinal circuit, such as place, grid, and border cells~\cite{moser2015place}.  Place cells  fire selectively at one or few locations in the environment. Place, grid and border cells
likely to interact with each other to yield a global
representation of the individual’s changing position.
Once encoded, the memories must be consolidated. Spatial memories, as other memories,  are thought to
be slowly induced in the neocortex by a  gradual
recruitment of neocortical memory circuits in
long-term storage of hippocampal memories~\cite{mcclelland1995there,squire1995retrograde,frankland2001alpha,moser2015place}.

The fast implementation of weight patterns in the hippocampal area is discussed under the term
\emph{synaptic consolidation} and  occurs within  minutes to hours, and as such is considered the ``fast'' type of consolidation.

According to our theory, the hippcampus would need to be well connected to the association areas of the cortex. Indeed, the hippocampus receives inputs from the unimodal and polymodal association areas of the cortex (visual, auditory, somatosensory) by a pathway involving the perirhinal and parahippocampal cortices which project to  the entorhinal cortex   which then projects to the hippocampus.  All these structures are part of  the MTL. The perirhinal and parahippocampal cortices also project back to the association areas of the cortex~\cite{gazzaniga2004cognitive}.


Figure~\ref{fig:NN-bio} (bottom right) also indicates a slow transfer  to long-term episodic memory.   The hypothesis is that  the index neurons and their latent representation form the basis for episodic memory! Biologically, this is referred to as \emph{system consolidation}, where hippocampus-dependent memories become independent of the hippocampus over a period of weeks to years. According to the standard model of  memory consolidation~\cite{squire1995retrograde,frankland2005organization} memory is retained in the hippocampus for up to one week after initial learning, representing the hippocampus-dependent stage. Later the hippocampus’ representations of this information become active in explicit (conscious) recall or implicit (unconscious) recall like in sleep. During this stage the hippocampus is ``teaching'' the cortex more and more about the information and when the information is recalled it strengthens the cortico-cortical connection thus making the memory hippocampus-independent.  Therefore from one week and beyond the initial training experience, the memory is slowly transferred to the neo-cortex where it becomes permanently stored. In this sense the MTL would act as a relay station for the various perceptual input that make up a memory and stores it as a whole event. After this has occurred the MTL directs information towards the neocortex to provide a permanent representation of the memory.

In our technical model we consider  two mechanisms for the transfer: Index neurons generated in the hippocampus and their representation pattern  might become part of the episodic memory, or neurons in the episodic memory are trained by replay:  this teaching process would be performed by the activation of the time index neurons, which then activate the ``sketchpad'' $\mathbf{a}_{e_t}$ which then trains the weight patterns of  time index neurons in long-term episodic memory.

As events are transferred from the hippocampus to episodic memory,  index neurons for places and entities and their latent representations would be consolidated in semantic long-term memory.

 The frontal cortex, associated with higher functionalities,  plays a role in which  new information  gets encoded as episodic and semantic memory and what gets forgotten~\cite{gluck2013learning}.


 The consolidation of  memory might be guided by novelty, attention, and emotional significance.
{ There is growing evidence that the amygdala is instrumental for storing  emotionally significant memories. The amygdala belongs to  the MTL and consists of several nuclei but is not considered to be a part of  memory itself~\cite{cahill1995amygdala}. } {The amygdala and the orbitofrontal cortex might also provide reward-related information to the hippocampus~\cite{rolls2010computational}. }

It has been shown in many studies that a loss of function of the hippocampus/MTL brain region leads to a loss of the consolidation of memory  to episodic long-term memory, but that this loss does not affect semantic memory. Our model supports this hypothesis, since semantic memory only relies on the latent representation of subject, predicate, and object, whereas episodic memory also relies on a latent representation of time, i.e., $\mathbf{a}_{e_t}$.


\subsection{Tensor Memory Hypothesis} 

The hypothesis states that  semantic memory and episodic memory are  implemented as  functions applied to the latent representations involved in the generalized entities  which include entities, predicates, and time.  Thus neither the knowledge graph nor the tensors ever needs to be stored explicitly! Due to the similarity to tensor decomposition, we call this the \textit{tensor memory hypothesis}.

\subsection{The Semantic Decoding Hypothesis and Association} 
\label{ssec:assoc}

$\mathbf{h}_t^{\textit{time}}$ is generated from sensory input and is the basis for episodic memory. For a semantic interpretation of sensory input and for a recall of episodic memory, $\mathbf{h}_t^{\textit{time}}$  can be rapidly decoded  by the semantic decoder shown in the center of Figure~\ref{fig:NN-bio}. As discussed in Sections~\ref{sec:senssem} and~\ref{sec:qa}, our model suggests that decoding happens by
the generation of $(s, p, o)$-triples by a stochastic sampling procedure.
Since a sensory input, in general, is described by several triples, this generation process is repeated several times,  generating a number of $(s, p, o)$-triples.  By sequential sampling, only one triples is active at a time and the ensemble of triples represents the query answer. Sequential sampling might also be influenced  by
 attention mechanisms, e.g., in the decoding of complex scenes~\cite{xu2015show,vinyals2015show,karpathy2015deep}.

{  The proposed model can be related to encoder-decoder networks~\cite{sutskever2014sequence} which produce text sequences,  whereas we produce a set of  likely triples.  $\mathbf{f}^M( \cdot)$ would be the encoder, potentially with internal feedback loops, $\mathbf{h}_t^{\textit{time}}$ would be the representation shared between encoder and decoder,  and the semantic decoder in our proposed model would correspond to the decoder. }

A clear indication that a semantic decoding is happening quickly is that an individual can describe a scene verbally immediately  after it has happened.\footnote{The language considered here is very simple and consists of triple statements.}

In the past,  a number of neural   winner-takes-all networks have been proposed where the neuron with the largest activation wins over all other neurons, which  are driven to inactivity~\cite{maass2000computational,hinton2006reducing}. Due to the inherent noise in real  spiking neurons, it is likely that  winner-takes-all networks  select one  of the neurons with large activities, not necessarily the one with the largest activity. Thus winner-takes-all sampling  might  be close to the sampling process specified in the theoretical model. {One might speculate that a winner-tales-all operation is performed in the complex formed by the dentate gyrus and the region III of hippocampus proper (CA3). It is known that CA3 contains many feedback connections, essential for winner-takes-all computations~\cite{marr1971simple,gluck2003computational,rolls2010computational}. CA3 is sometimes modelled as a continuous attractor neural network (CANN) with excitatory recurrent colateral connections and global inhibition~\cite{rolls2010computational}. }

The sampling denoises the scene interpretation. Each $(s, p, o)$-sample represents a sharp hypothesis; an advantage of the sampling approach is that no complex feedback mechanisms are required for the generation of  attractors, as in other approaches.

The proposed sampling procedure is a step-wise procedure which generates independent samples. An alternative might be a Gibbs sampler which could be implemented as easily. The advantage of a Gibbs sampler is that it does not require marginalization; a disadvantage is that the generated samples are not independent.   On the other hand, correlated samples might be the basis for free recall, associative thinking and chaining.

For association  we  can fix an entity $s$, generate its latent representation $\mathbf{a}_{e_s}$ and then sample a new entity $s'$ based on this latent representation, thus,  we can explore entities that are very similar to the original entity. Thus \textit{Barack Obama} might produce \textit{Michelle Obama}.  During sampling the roles of subjects might be interchanged. Thus  the triple \textit{(Obama, presidentof,  USA)} might produce samples describing properties and relationships of the USA.


The restricted Boltzmann machine (RBM) might be an interesting option for supporting the decoding process~\cite{smolensky1984harmony,hinton2010practical}.

As discussed in the caption of Figure~\ref{fig:NN-bio} it is even possible to operate the model in reverse: If we consider a person $s$ to be the input, marginalize out $p$ and $o$ and consider $\mathbf{h}_t^{\textit{time}}$ as the output, then we can recall when we met the person  by exciting the time index neuron, and we can even recall her appearance
 by operating $\mathbf{f}^M(\cdot)$ in reverse.

According to our model the recall of episodic memory would be driven by an activation of the time latent representation $\mathbf{a}_{e_t}$, which is then semantically decoded and elucidates sensory impressions. This fits the subjective feeling of a  reconstruction of past memory.

%
%


$M$-mapping, prediction, and semantic decoding are fast operations possibly involving many parts of the cortex.\footnote{The physicist Eugene Wigner  has speculated on the  ``The Unreasonable Effectiveness of Mathematics in the Natural Sciences''~\cite{wigner1960unreasonable}; in other words mathematics is the right code for the natural sciences. Similarly, semantics might be considered the language for the world, in as far as humans are involved and one might speculate about its unreasonable effectiveness as well.}

{The semantic coding and decoding in our proposed  model  might biologically  be located in the MTL.  There is growing evidence that the hippocampus plays an important role not just in  encoding but also in  decoding of memory and is involved in the retrieval of information from long-term memory~\cite{gazzaniga2004cognitive}.  The binding of items and cortex (BIC) theory states that   the perirhinal cortex (anterior part of the parahippocampal region) connects to the ``who'' and  what'' pathways of unimodal sensory  brain regions. In our model this information is  decoded into $(s, p, o)$-triples. In contrast the  ``when''  and  ``where''  parts pass through the posterior part of the  parahippocampal region.
Both types of information then pass  through the entorhinal cortex but only converge within the hippocampus where it  enables a full recognition of an episodic event~\cite{eichenbaum2007medial,diana2007imaging,ranganath2010binding,gazzaniga2004cognitive}.   The ``what'' pathway is involved in the  anterior temporal (AT) system also involving parts of the temporal lobe (ventral temporopolar cortex) and is associated with semantic memory.  The  ``where'' pathway is part of the posterior medial (PM) system also involving parts of the parietal cortex (retrospinal cortex) and is associated with semantic memory.   }

\subsection{Semantic Memory and Episodic Memory} 
\label{sec:epissem}

As discussed, episodic memory is implemented in form of time index neurons and their latent
 representations $\mathbf{a}_{e_t}$,  and is decoded using the latent representations for subjects, predicates and objects. But what about semantic memory?  In Section~\ref{sec:modelquery} (Figures~\ref{fig:ffNNx} and Figures~\ref{fig:ffNNy}) we describe a semantic memory which is implemented as a separate indicator mapping function  that is also based on the latent representations of subject, predicate and object.

Biologically it might be quite challenging to transfer episodic memory into semantic memory.  An alternative, with a number of interesting consequences, is that the semantic memory is generated from episodic memory by marginalizing time, as shown in the bottom of Figure~\ref{fig:SensorChain}.
 In this interpretation, semantic memory is a long-term storage for episodic memory.
 Thus to  answer the query ``what events happened at time $t$'', the system needs to retrieve $\mathbf{a}_{e_t}$ and perform a semantic decoding into $(s, p, o)$-triples. In contrast, to decode a triple from semantic memory,  $\mathbf{a}_{e_t}$ is replaced with $\mathbf{\bar{{a}}} = \sum_t {\mathbf{a}}_{e_t}$, which can either be calculated by inputting a vectors of ones or by learning a long-term average (Figure~\ref{fig:SensorCh}(D)).\footnote{One can also easily be only considering semantic memory of a certain time span  by just inputting ones for the time index neurons of interest.}

This form of a semantic memory is very attractive since it requires no additional modelling effort and can use the same structures that are needed  for episodic memory! It has been argued that  semantic memory is information we have encountered repeatedly, so often that the actual learning episodes are blurred~\cite{conway2009episodic,gluck2013learning}.
 A gradual transition from episodic to semantic memory can take place, in which episodic memory reduces its sensitivity and association to particular events, so that the information can be generalized as semantic memory.
  Without doubt, semantic and episodic memories support one another~\cite{greenberg2010interdependence}.
  Thus  some theories speculate that  episodic memory may be the
``gateway'' to semantic memory~\cite{baddeley1974working,squire1987memory,baddeley1988cognitive,steyvers2004word,socher2009bayesian,mcclelland1995there,yee2014cognitive,kumar2015ask}.
 \cite{morton2013interactions} is a recent overview on the topic.
Our model would also support the alternative view of Tulving that episodic memory depends on the semantic memory, i.e., the representations of entities and predicates~\cite{tulving2002episodic,gluck2013learning}.
{But note that studies have also found an independent formation of semantic memories, in case that the episodic memory is dysfunctional,  as in certain amnesic patients: Amnesic patients might learn new facts without remembering the episodes during which they have learned the information~\cite{gazzaniga2004cognitive}. This phenomenon is supported by our proposed model  since there is a direct path from sensory input to the representations of subject, predicate and object.}


Our model  supports inductive  inference in form of a probabilistic materialization.
Certainly humans are   capable of some form of logical inference, but this might be   a faculty of working memory. The approximations that are performed in the tensor models, respectively in the the multiway neural networks, lead to a form of a probabilistic materialization, or unconscious inference:   As an example,  consider that we know that Max lives in Munich. The probabilistic materialization that happens in the factorization should already predict that Max also lives in Bavaria and in Germany. Thus both facts and  inductively inferred facts about an entity are represented in its  local environment. {
There is a certain danger in probabilistic materialization, since it might lead to  overgeneralizations, reaching  from national prejudice to false memories.   In fact in many studies it has been shown that individuals produce false memories but  are personally absolutely  convinced of their truthfulness~\cite{roediger1995creating,loftus1996myth}.
}

{Our model assumes symmetrical connections between index neurons and representation neurons.  The biological plausibility of symmetric weights has been discussed intensely in computational neuroscience and many biologically oriented models have that property~\cite{hopfield1982neural,hinton2006reducing}. Reciprocal connectivity is abundant in the brain, but perfect symmetry is typically not observed.
}




%
%
%
%


%


\subsection{Online Learning and the  Semantic-Attractor Learning Hypothesis}

An interesting feature of the proposed model is that no learning or adaptation is necessary in operation, as long as sensory information can be described by the entities and predicates already known. The only structural adaptation that happens online is the forming of the index neuron $e_t$ and its representation pattern $\mathbf{a}_{e_t}$.

If decoding is not successful, e.g., if the decoded triples have low likelihood, one might consider  a mechanism for introducing new index neurons with new  latent representations for entities and predicates not yet stored in memory.
 Thus, only when the available resources (entities and predicates) are insufficient for  explaining the sensory data, new index neurons for entities and predicates are introduced.

At a slower time scale it might  be necessary to fine-tune all parameters in the system, possibly also the latent representations for entities and predicates. One might look at the model in Figure~\ref{fig:NN-bio} as a complex neural network with inputs $u_{:, :, t}$ and targets $(s, p, o)$, possibly with some recurrence via  the prediction module. Powerful learning algorithms are available to train such a system in a supervised way, and this might be the solution in a technical application.   Of course for a biological system, the target information is unavailable.

So how can such a complex system be trained without clear target information?  The future prediction model can be trained to lead to
 high quality predictions of future sensory inputs \cite{hinton1994autoencoders,dayan1995helmholtz,rao1999predictive,knill2004bayesian,kording2004bayesian,tenenbaum2006theory,griffiths2008bayesian,george2009towards,friston2010free}.
 For the remaining parameters we suggest a form of bootstrap learning: the model parameters should be adapted such that they lead to stable semantic interpretation of sensory input.    We call this the \textit{semantic-attractor learning hypothesis}:  In a sense the semantic descriptions form attractors for decoded  sensory data and, conversely,  the attractors are adapted based on sensory data.  This can be related to the phenomenon of  ``emergence'' which is a process whereby larger  patterns and regularities arise through interactions among smaller or simpler entities that themselves do not exhibit such properties. Thus the \textit{emerging semantics hypothesis} is that the semantic description is an emergent property of the sensory inputs!

\subsection{Working Memory Exploits the Memory Representations for Tasks like Prediction and Decision Making}
\label{sec:wm2}

On the top right of Figure~\ref{fig:NN-bio} we see a future-prediction model which estimates the next $\mathbf{h}_t^{\textit{time}}= \mathbf{a}_{e_t}$ based on its past values and based on the latent representation for the individual $\mathbf{a}_{e_s}$.
Note that $\mathbf{a}_{e_s}$ is not considered constant; for example, an individual might be diagnosed with a disease, which would be reflected in a change in  $\mathbf{a}_{e_s}$.  Large differences between predicted and sensory-decoded latent representations $\mathbf{a}_{e_t}$ represent novelty and might be a component of an  attention mechanism. As discussed before, novelty might be an important factor that determines which sensory information is stored in episodic memory, as speculated by other models and supported by cognitive studies~\cite{dayan2000learning,itti2005bayesian,schmidhuber2009driven,friston2010free,barto2013novelty}.

An interesting aspect is that the predicted $\mathbf{h}_t^{\textit{time}}$ can be semantically decoded for a cognitive analysis of predicted events (see Figure~\ref{fig:NN-bio}) and can lead to mental imagery, a sensory representation of predicted events. Mental imagery   can be viewed as the conscious and explicit manipulation of simulations in working memory to predict future events~\cite{barsalou2009simulation}. The link between episodic  memory and  mental imagery has been studied in~\cite{schacter2012future}  and ~\cite{hassabis2007deconstructing}.


In Section~\ref{sec:predict} we discussed a  predictive ARX model and an RNN model. In human cognition, both might be significant: The RNN would be part of  the model dynamics, whereas the ARX model would  purely serve as a predictive component.



 Prediction of events and actions on a semantic level is sometimes considered to be one of the important functions of a cognitive working memory~\cite{o2006making}. Working memory is the limited-capacity store for retaining information over the short term and for performing mental operations on the contents of this store. As in our prediction model, the contents of working memory could either originate from sensory input, the episodic buffer,  or  from semantic memory~\cite{gazzaniga2004cognitive}.
Cognitive models of working memory are described in~\cite{baddeley1974working,baddeley1992working, baddeley2012working,cowan1997attention,ericsson1995long} and  computational  models are described in  \cite{mozer1993neural,durstewitz2000neurocomputational,frank2001interactions,burgess2005computational,jonides2008mind,o2006making}.

The terms ``predictive brain''  and ``anticipating brain'' emphasize the importance of ``looking into the future'', namely prediction, preparation,
anticipation, prospection or expectations in various cognitive domains~\cite{clark2013whatever}.
Prediction has been a central concept in recent trends in computational neuroscience, in particular in recent Bayesian approaches to brain modelling~\cite{hinton1994autoencoders,dayan1995helmholtz,rao1999predictive,knill2004bayesian,kording2004bayesian,tenenbaum2006theory,griffiths2008bayesian,george2009towards,friston2010free}. In some of these approaches, probabilistic generative models generate hypothesis about observations (top-down) assuming hidden causes,  which are then aligned with actual observations (bottom-up).

Working memory is not the only brain structure involved in prediction.  Predictive control is crucial for fast and ballistic movements where the cerebellum plays a crucial role in implicit tasks. {The cerebellum is involved in trial-and-error learning based on predictive error signals~\cite{gazzaniga2004cognitive}.
Reward prediction is a task of the basal ganglia where dopamine neurons encode both present rewards and future rewards, as a basis for reinforcement learning~\cite{gazzaniga2004cognitive,gluck2013learning}.}

Working memory, assumed to be located in the frontal cortex, can use the representations in Figure~\ref{fig:NN-bio}  in many ways, not just for prediction. In general, working memory is closely tied to complex problem solving, planning, organizing,  and  decision support, and might assume an important role in consciousness.  There is evidence that a strong working memory is associated with general intelligence~\cite{gluck2013learning}.

{One influential cognitive model  of working memory is Baddeley’s multicomponent model~\cite{baddeley1974working}. Cognitive control is executed by a  central executive system. It is supported by two subsystems responsible for maintenance and rehearsal: the phonological loop, which maintains verbal information and
the visuospatial sketchpad, which maintains visual and spatial information.
More recently the
episodic buffer has been added to the model. The episodic buffer  integrates short-term and long-term memory, holding and manipulating a limited
amount of information from multiple domains in time and spatially sequenced episodes (Figure~\ref{fig:mem}).
There is an emerging consensus that functions of working memory are located in the prefrontal cortex and that a number of other brain areas are recruited~\cite{o199911,gazzaniga2004cognitive}.}  {More precisely, the central executive is attributed to the dorsolateral prefrontal cortex, the phonological loop with
the left ventrolateral prefrontal cortex (the semantic information is anterior  to  the phonological information) and the  visuospatial sketchpad in the right  ventrolateral prefrontal cortex~\cite{gluck2013learning}.  }
The function of the frontal lobe, in particular of the orbitofrontal cortex,   includes the ability to project future consequences (predictions) resulting from current actions~\cite{gluck2013learning}.

\section{Conclusions and Discussion}
\label{sec:concl}

We have discussed how a number of technical memory functions can  be realized by  representation learning and we have made the connection to  human memory.
A key assumption is that a knowledge graph does not need to be stored explicitly, but only latent representations of  generalized entities need to be stored from which the knowledge graph can be reconstructed and inductive inference can be performed (tensor memory hypothesis).  Thus,   in contrast to the knowledge graph, where an entity is represented by a single node in a graph and its links, in embedding learning,   an entity has a distributed representation in form of a latent vector, i.e.,  in form of multiple  latent components. Unique representations  lead to a global propagation of information across all memory functions during  learning~\cite{nickel_three-way_2011}.

We proposed   that the  latent representation for a time $t$,
which summarizes all sensory information present at time $t$, is the basis for episodic memory and that semantic memory depends on the latent representations of subject, predicate, and object. One theory we support is that semantic memory is  a long-term aggregation of episodic memory.
 The full episodic experience depends on both semantic (``who'' and ``what'') and context representations (``where'' and ``when''). On the other hand there is also a certain independence: the pure storage of episodic memory does not depend on semantic memory and semantic memory can be acquired even without a functioning episodic memory.  The same relationships between semantic and episodic memories can be found  in the human brain.

The latent representations of the  semantic memory, episodic memory, and sensory memory    can support working memory functions like prediction and decision support. In addition to the latent representations, the models contain parameters (e.g., neural network weights) in mapping functions,  memory models and   prediction models.  One can make a link between   those parameters and implicit skill memory~\cite{schacter1987implicit}. Refining the mapping from sensory input to its latent representation corresponds to perceptual learning in cognition.

We showed how  both a recall of previous memories and the mental imagery of future events and sensory impressions can be supported by the presented model.

More details on concrete technical solutions can be found in~\cite{Crist2015,EstebanEvent} where we also present successful applications to clinical decision modeling, sensor network modeling and recommendation engines.

{%
	\bibliography{LearnMemEmb}
	\bibliographystyle{plainnat}
}

\section{Appendix}

\subsection{Cost Functions}

The cost function is the sum of several terms. The tilde notation $\mathcal{\tilde{X}} $ indicates subsets which correspond to  the facts known in training. If only positive  facts with $\textit{Value }= \textit{True}$ are known, negative facts can  be generated using,  e.g., local closed world assumptions~\cite{nickel2015}. We use negative log-likelihood cost terms. For a Bernoulli likelihood,
$- \log P(x| \theta) = \log [1 + \exp\{(1-2x)\theta\}]$ (cross-entropy) and for a Gaussian likelihood
$- \log P(x| \theta) = const +  \frac{1}{2 \sigma^2}(x-\theta)^2$.

\subsubsection{Semantic KG Model}

The cost term for the semantic KG model is
\[
\textrm{cost}^{\textit{semantic}} = - \sum_{ x_{s, p, o} \in \mathcal{\tilde{X}} }  \log P(x_{s, p, o} |
\theta^{\textit{semantic}}_{s, p, o} (A, W))
\]
 where $A$ stands for the latent representations and $W$ stands for the parameters in the functional mapping.

\subsubsection{Episodic Event Model}
\[
\textrm{cost}^{\textit{episodic}} =
- \sum_{z_{s, p, o, t} \in \mathcal{\tilde{Z}} }  \log P(z_{s, p, o, t} |
\theta^{\textit{episodic}}_{s, p, o, t} (A, W))
\]

\subsubsection{Sensory Buffer}

\[
\textrm{cost}^{\textit{sensory}} =
-   \sum_{  u_{q, \gamma, t} \in \mathcal{\tilde{U}} }  \log P(u_{q, \gamma, t} |
\theta^{\textit{sensory}}_{q,  \gamma, t} (A, W))
\]

\subsubsection{Future-Prediction  Model}

The cost function for the ARX prediction model is
\[
\textrm{cost}^{\textit{predict}} =
-  \sum_{t}   \log P(\mathbf{a}_{e_{t}} |
\mathbf{f}^{\textit{predict}}(\mathbf{a}_{e_{t-1}} , \mathbf{a}_{e_{t-2}} , \ldots, \mathbf{a}_{e_{t-W}} , \mathbf{a}_{e_{\textit{indiviual}}}, A, W)
\]


\subsubsection{Regularizer}

To regularize the solution we add
\[
\lambda_{A} \| A \|_F^2 + \lambda_{W} \| W \|_F^2
\]
where $\| \cdot \|_F$ is the Frobenious norm and  where $\lambda_{A}\ge 0$ and  $\lambda_{W}\ge 0$  are regularization parameters.  If we use $M$-mappings, we regularize $M$ instead of $A$ and we include $\lambda_{M} \| M \|_F^2$.

\subsection{Sampling using Function Approximators}
\label{sec:sfa}

\begin{figure}
	\centering
	\includegraphics[width=\columnwidth]{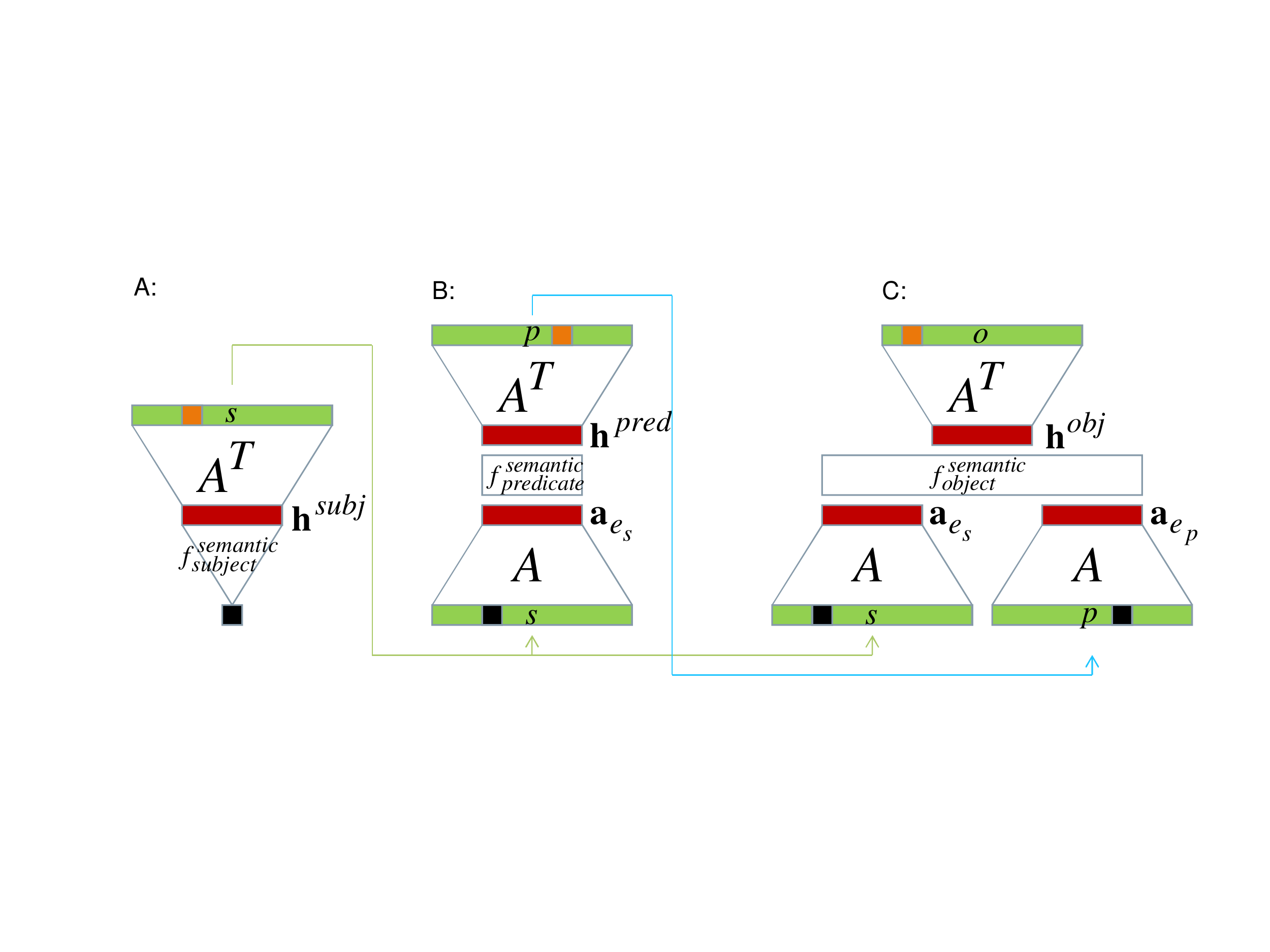}
	\caption{%
Semantic KG sampling using a general function approximator, e.g., a feedforward neural network.
A: A subject is sampled based on $P(s) \propto \exp \beta \mathbf{a}^{\top}_{e_s } \mathbf{h}^{\textit{subject}}$.
$\mathbf{h}^{\textit{subject}}$ is a learned latent vector.
B: An predicate is sampled based on $P(p|s) \propto \exp \beta \mathbf{a}^{\top}_{e_p } \mathbf{h}^{\textit{predicate}}$.
$\mathbf{h}^{\textit{predicate}}$ is  a learned function of the sample  $s$.
C: An object is sampled based on $P(o|s, p) \propto \exp \beta \mathbf{a}^{\top}_{e_o } \mathbf{h}^{\textit{object}}$.
$\mathbf{h}^{\textit{object}}$ is  a learned function of the sample  $s, p$.
}
	\label{fig:funcspo}
\end{figure}

\begin{figure}
	\centering
	\includegraphics[width=\columnwidth]{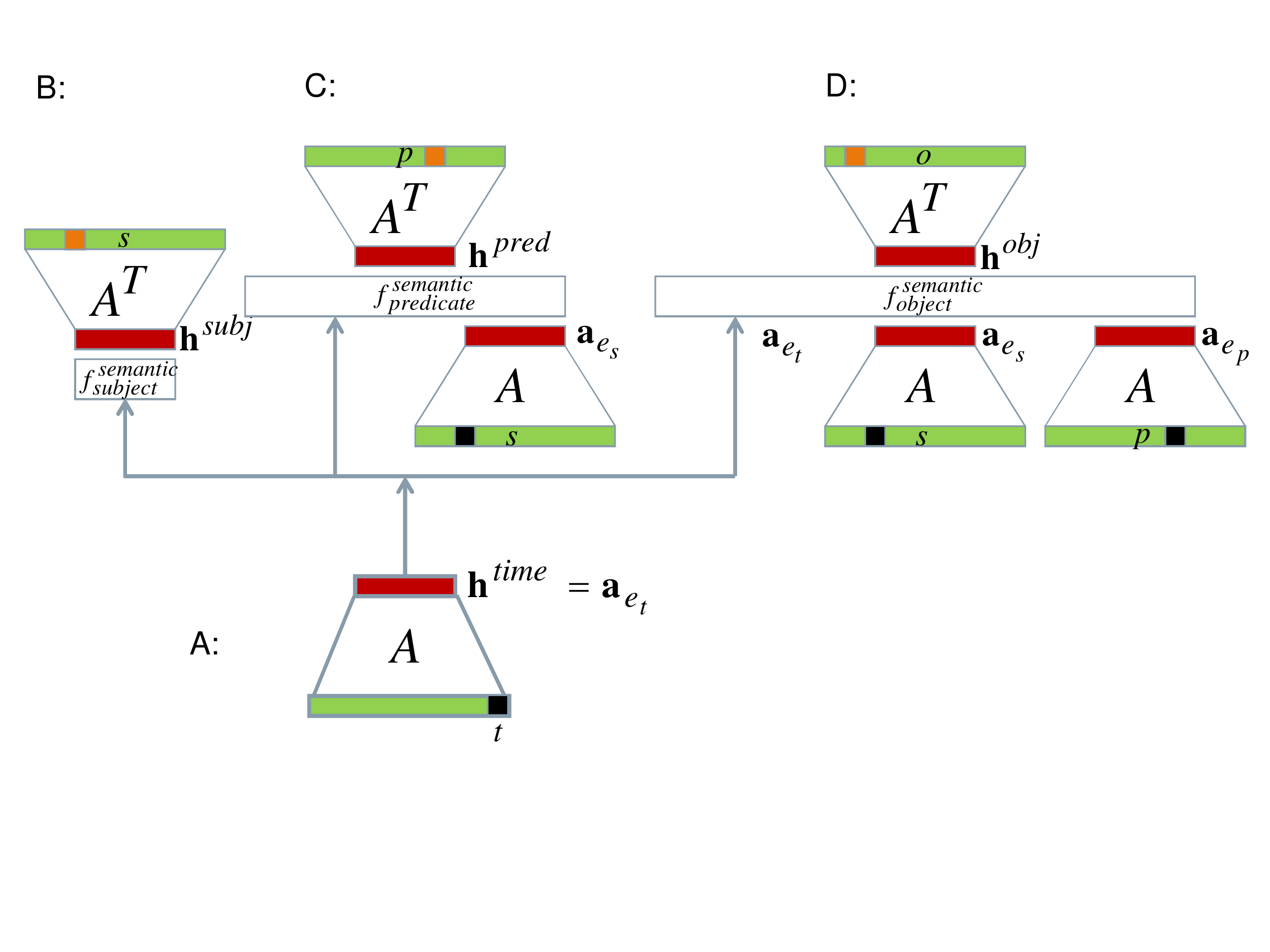}
	\caption{%
 The semantic decoding  using a general function approximator, e.g., a feedforward neural network.  A:  The sensory memory produces $\mathbf{h}_t^{\textit{time}} = \mathbf{ a}_{e_t}$ based on $u_{:, :, t}$. $\mathbf{ a}_{e_t}$ is represented in the weights of  index neuron $e_t$. B: $\mathbf{ a}_{e_t}$ is then the input to the left model and  a subject $s$ is sampled based on
  $P(s|t) \propto \exp \beta \mathbf{a}^{\top}_{e_s } \mathbf{h}^{\textit{subject}}$.
  C:  With $ \mathbf{ a}_{e_t}$ and the sampled subject as inputs,  a predicate $p$ is sampled based on $P(p|s, t) \propto \exp \beta \mathbf{a}^{\top}_{e_p } \mathbf{h}^{\textit{predicate}}$.
  D:  With $\mathbf{  a}_{e_t}$ and the sampled subject and predicate as inputs,  an object $o$ is sampled based on $P(o|s, p, t) \propto \exp \beta \mathbf{a}^{\top}_{e_o } \mathbf{h}^{\textit{object}}$.
}
	\label{fig:SensorCh}
\end{figure}

Figure~\ref{fig:funcspo} shows how samples using function approximators (e.g., a NN) can be generated for the semantic KG and Figure~\ref{fig:SensorCh} shows the semantic decoding.

\end{document}